\documentclass{article}

\usepackage[preprint]{neurips_2026}

\usepackage[utf8]{inputenc} % allow utf-8 input
\usepackage[T1]{fontenc}    % use 8-bit T1 fonts
\usepackage{url}            % simple URL typesetting
\usepackage{booktabs}       % professional-quality tables
\usepackage{amsfonts}       % blackboard math symbols
\usepackage{nicefrac}       % compact symbols for 1/2, etc.
\usepackage{microtype}      % microtypography
\usepackage{multirow}
\usepackage{graphicx}
\usepackage{geometry}
\usepackage{color}
\usepackage{amsmath}
\usepackage{enumitem}
\usepackage[table]{xcolor}
\definecolor{promptblue}{RGB}{25, 200, 200}

\usepackage{svg}
\usepackage{caption}
\usepackage{utfsym}
\usepackage{colortbl}
\usepackage{xcolor}
\usepackage{makecell}
\usepackage{wrapfig}
\usepackage{array}
\usepackage{subcaption}
\usepackage[colorlinks,linkcolor=blue]{hyperref}

\usepackage{fontawesome5}

\definecolor{baserow}{RGB}{235, 242, 250}
\definecolor{allrow}{RGB}{220, 240, 220}

\title{SpaceDG: Benchmarking Spatial Intelligence under Visual Degradation}
\author{%
    Xiaolong Zhou$^{2,3}$\hspace{0.6em}
    Yifei Liu$^{2,6}$\hspace{0.6em}
    Ziyang Gong$^{1}$\hspace{0.6em}
    Jiarui Li$^{4}$\hspace{0.6em}
    Qiyue Zhao$^{4}$\hspace{0.6em}
    Muyao Niu$^{5}$ \\
    \textbf{Yuanyuan Gao}$^{2,7}$\hspace{0.6em}
    \textbf{Le Ma}$^{2}$\hspace{0.6em}
    \textbf{Xue Yang}$^{1}$\hspace{0.6em}
    \textbf{Hongjie Zhang}$^{2}$\hspace{0.6em}
    \textbf{Zhihang Zhong}$^{1,\dagger}$ \\ [0.35em]
    $^{1}$Shanghai Jiao Tong University \qquad
    $^{2}$Shanghai Artificial Intelligence Laboratory \\
    $^{3}$University of Electronic Science and Technology of China \qquad
    $^{4}$Chongqing University \\
    $^{5}$The University of Tokyo \qquad
    $^{6}$Beihang University \qquad
    $^{7}$Northwestern Polytechnical University \\ [0.3em]
    $^{\dagger}$Corresponding author \hspace{0.4em}
    \faGithub \hspace{0.2em} \href{https://github.com/Visionary-Laboratory/SpaceDG}{https://github.com/Visionary-Laboratory/SpaceDG}
}

\begin{document}
\maketitle

\begin{figure*}[h]
\centering
\includegraphics[width=1\linewidth]{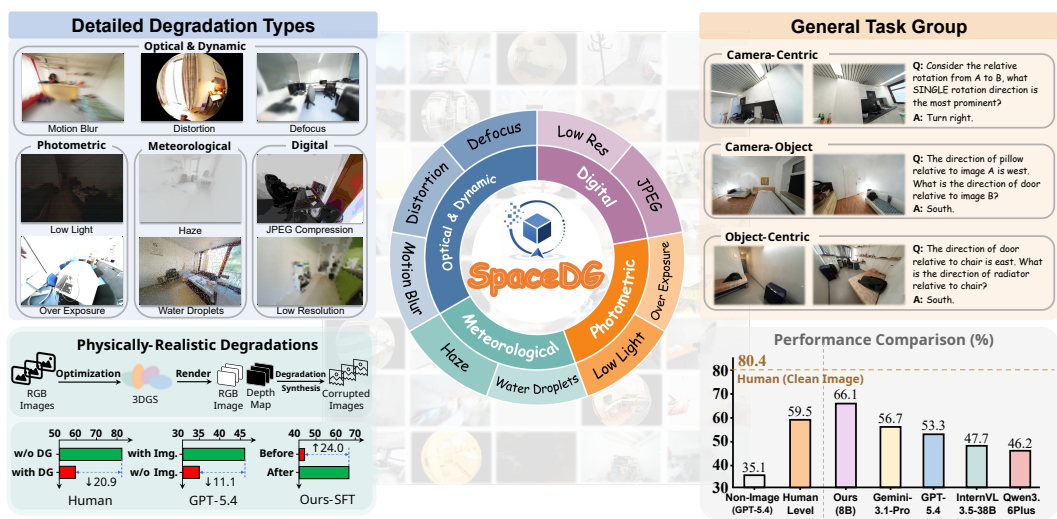}
\caption{Overview of SpaceDG. \textbf{Top-left:} Nine physically-realistic degradations across four categories. \textbf{Top-right:} Three spatial task groups covering camera-centric, camera-object, and object-centric questions. \textbf{Mid-left:} 3DGS-based degradation data engine. \textbf{Bottom-right:} Performance comparison of representative models against human-on-clean-image and non-image baselines, reflecting the upper bound and lower bound respectively. \textbf{Bottom-left:} additional performance comparisons.}
% \vspace{-0.3cm}
\label{fig:teaser}
\end{figure*}

\begin{abstract}
Multimodal Large Language Models (MLLMs) have made rapid progress in spatial intelligence, yet existing spatial reasoning benchmarks largely assume pristine visual inputs and overlook the degradations that commonly occur in real-world deployment, such as motion blur, low light, adverse weather, lens distortion, and compression artifacts. This raises a fundamental question: how robust is the spatial intelligence of current MLLMs when visual observations are imperfect? To answer this question, we introduce \textbf{SpaceDG}, the first large-scale dataset for degradation-aware spatial understanding. It is constructed with a physically grounded degradation synthesis engine that embeds degradation formation process into 3D Gaussian Splatting (3DGS) rendering, enabling realistic simulation of nine degradation types. The resulting dataset contains approximately 1M QA pairs with over 160K images. We further introduce \textbf{SpaceDG-Bench}, a human-verified benchmark with 1,102 unique questions spanning 11 reasoning categories and 9 visual degradation types, yielding 10K VQA instances. Evaluating 25 open- and closed-source MLLMs reveals that visual degradations consistently and substantially impair spatial reasoning, exposing a critical robustness gap. Finally, we show that finetuning on SpaceDG markedly improves degradation robustness and can even surpass human performance under degraded conditions without any performance drop on clean images, highlighting the promise of degradation-aware training for robust spatial intelligence.
\end{abstract}

\section{Introduction}
Multimodal Large Language Models (MLLMs) have achieved remarkable success in spatial intelligence, bridging the crucial gap between 2D visual recognition and 3D physical reasoning~\cite{liu2023llava, wu2025spatialmllmboostingmllmcapabilities, black2026pi0visionlanguageactionflowmodel}. As a fundamental capability of visual cognition~\cite{yang2024think,luo2025visual,li2025robotic}, spatial intelligence poses immense challenges to a model's ability to perceive, parse, and reason within the complex real world. To evaluate and advance this, researchers have proposed a myriad of benchmarks~\cite{yang2025mmsi, zhang2025dsibenchbenchmarkdynamicspatial, yang2025cambrians, jia2026omnispatialcomprehensivespatialreasoning, wang2026mindcubespatialmentalmodeling, li2025viewspatialbenchevaluatingmultiperspectivespatial,yang2026stepping}, upon which current state-of-the-art models~\cite{Chen_2024_CVPR, wu2025spatialmllmboostingmllmcapabilities, vst, yang2025cambrians, sensenova-si} demonstrate impressive spatial awareness, positioning them as the foundational brains for embodied agents and autonomous systems.

Existing spatial benchmarks predominantly evaluate MLLMs under a ``perfect observation'' assumption, using clean, high-resolution, and well-illuminated images. 
Yet in real-world embodied and autonomous systems, visual observations are produced by imperfect sensing pipelines, where degradations naturally arise during acquisition, transmission, and deployment. 
These degradations are not merely artificial corruptions, but common conditions faced by agents operating in physical and resource-constrained environments. 
They have been extensively studied in low-level vision and computational photography, spanning motion blur~\cite{su2017deep,nah2017deep,zhong2020efficient,zhong2023real,zhong2023blur}, low-resolution imaging~\cite{dong2015image,ledig2017photo,wang2018esrgan,lu2022transformer}, geometric distortion~\cite{liu2020deep,zhong2021towards,cao2022learning}, low-light~\cite{chen2018learning,niu2023visibility,niu2023nir}, and adverse weather~\cite{he2010single,fu2017clearing}. 
Under such conditions, the robustness of spatial intelligence becomes a critical requirement, since spatial reasoning often depends on fine-grained geometric evidence, including object boundaries, relative positions, and multi-view consistency. 
Despite this rich literature on degradation and recent advances in benchmarking general VLM robustness~\cite{tang2025robustr1, saxena2026vlmrobustbenchcomprehensivebenchmarkrobustness}, how current MLLMs perform spatial reasoning under imperfect observations remains an open question.

To systematically answer this question, a suitable benchmark must satisfy three requirements: it should introduce realistic visual degradations, preserve the underlying 3D spatial structure, and support diverse spatial reasoning tasks with reliable ground truth. To fill this gap, we introduce SpaceDG and SpaceDG-Bench, the first large-scale VQA dataset and benchmark dedicated to degradation-aware spatial understanding, and conduct a comprehensive evaluation of current MLLMs under imperfect visual observations.

Specifically, we develop an automatic degradation data engine. First, we reconstruct multi-view images into geometrically accurate 3D Gaussian Splatting (3DGS)~\cite{kerbl3Dgaussians} representations and pair them with auto-annotated spatial QA\cite{gao2026holispatialevolvingvideostreams} . Second, on top of the reconstructed 3DGS, we design a physically realistic degradation synthesis pipeline that simulates nine representative degradations across four categories: (1) \textbf{optical and dynamic degradations}, including defocus, distortion and motion blur; (2)  \textbf{meteorological degradations}, including haze and water droplets; (3) \textbf{photometric degradations}, including low light and over-exposure; and (4) \textbf{digital degradations}, including JPEG compression and low resolution. Each degradation is generated from underlying physical formation process, as shown on the left in Figure~\ref{fig:teaser}.

Leveraging this engine, we construct SpaceDG, a large-scale dataset derived from nearly 1,000 scenes in ScanNet++~\cite{yeshwanth2023scannet++}. SpaceDG comprises approximately 1M QA pairs over more than 160K images and covers a diverse range of visual degradations. To establish a rigorous evaluation protocol, we further introduce SpaceDG-Bench, a manually curated and verified benchmark comprising 1K high-quality QA pairs. For comprehensive assessment, we systematically design 11 distinct question categories, encompassing camera-centric, object-centric and object-camera relation questions with single-view or multi-view images.

We conduct a comprehensive evaluation of 25 models and identify four key findings. \textbf{First}, visual degradations consistently impair spatial reasoning across all evaluated MLLMs,  highlighting the need for degradation-aware spatial evaluation. \textbf{Second}, humans also suffer clear performance drops under degraded conditions. This suggests that the design of MLLMs should not simply imitate human perception, but should learn degradation-aware spatial knowledge to better handle diverse real-world visual inputs. \textbf{Third}, degradation-based SFT yields substantial improvements on both clean and degraded inputs, indicating that exposure to physically grounded degradations can enhance robust spatial understanding. \textbf{Finally}, we observe that visual degradations affect fine-grained object-level perception, such as object counting, more strongly than certain geometric reasoning tasks, such as camera-centric translation, revealing that detailed visual grounding is particularly sensitive to degraded visual evidence. 

\section{Related works}

\paragraph{Spatial intelligence of MLLMs} Recent advances in spatial MLLMs have expanded their capabilities from basic visual understanding~\cite{qwen3.5, wang2025internvl35advancingopensourcemultimodal, kimiteam2025kimivltechnicalreport, coreteam2025mimovltechnicalreport} to fine-grained spatial reasoning~\cite{yang2025cambrians, sensenova-si, wu2025spatialmllmboostingmllmcapabilities, vst, cheng2024spatialrgptgroundedspatialreasoning, daxberger2025mmspatialexploring3dspatial} with large-scale spatial datasets. For example, Cambrian-S~\cite{yang2025cambrians}, VST~\cite{vst}, and SenseNova-SI~\cite{sensenova-si} adopt VSI-590K, 4.1M samples, and SenseNova-SI-8M, respectively, to boost spatial intelligence. To evaluate these models, researchers have developed various benchmarks~\cite{yang2024think, yang2025mmsi, zhang2025dsibenchbenchmarkdynamicspatial, zhou2025vlm4d, wang2026mindcubespatialmentalmodeling}. However, both spatial models and benchmarks operate under a ``perfect image assumption'', where images are clear and well illuminated, failing to reflect physical constraints and visual imperfections in real-world deployment.

\paragraph{Robustness of MLLMs Against Visual Degradations} In unconstrained physical environments, visual inputs inevitably suffer from degradations caused by dynamic motion, adverse weather, and sensor limitations. Such corruptions have been standardized in ImageNet-C~\cite{hendrycks2019benchmarkingneuralnetworkrobustness}, and recent works have begun evaluating MLLM robustness against common image corruptions~\cite{cui2023robustnesslargemultimodalmodels, saxena2026vlmrobustbenchcomprehensivebenchmarkrobustness, usama2025analysingrobustnessvisionlanguagemodelscommon, tang2025robustr1, fan2025v2rbenchholisticallyevaluatinglvlm}. However, existing studies mainly focus on semantic understanding, object recognition, or basic visual reasoning~\cite{usama2025analysingrobustnessvisionlanguagemodelscommon, fan2025v2rbenchholisticallyevaluatinglvlm, tang2025robustr1}. The robustness of MLLMs under visual degradation for fine-grained spatial intelligence remains unclear. 

\paragraph{3DGS Representation and Data Synthesis}
3DGS~\cite{kerbl3Dgaussians} has rapidly emerged as an efficient and expressive 3D representation for real-time novel view synthesis and scene reconstruction. Recent work further improves its quality, scalability, and compactness from several perspectives, including more structured or expressive Gaussian formulations~\cite{scaffoldgs, octreegs, Yu2023MipSplatting, gao2025proxy, chen2024pgsr}, large-scale scene reconstruction~\cite{liu2025citygaussian, liu2024citygaussianv2, Gao_2025_ICCV, 10655941}, and 3DGS compression~\cite{lee2024c3dgs, Liu_2025_CVPR, fan2024lightgaussian}. In parallel, another line of research models realistic visual degradations, such as motion blur~\cite{9025509, zhao2024badgaussians, niu2026motionawareanimatablegaussianavatars}, defocus blur~\cite{Lee_2023_CVPR, wang2024dofgs}, low-light conditions~\cite{9878457, wei2021physics}, and geometric or optical distortions~\cite{liao2024fisheyegslightweightextensiblegaussian, wu20253dgut}. Motivated by these advances, we adopt 3DGS as a geometry-consistent and renderable scene representation, and couple it with degradation-specific physical formation models to synthesize realistic degraded observations while preserving the underlying 3D ground truth.

\section{SpaceDG}
This section presents SpaceDG and SpaceDG-Bench, the first dataset and benchmark for spatial intelligence under visual degradations. We introduce our proposed data engine, starting with 3DGS-based scene representation and QA initialization, followed by a physically realistic degradation synthesis pipeline. Then we detail the constructed dataset, covering diverse spatial tasks, multiple viewpoints, and various visual degradations.

\begin{figure*}[t]
\centering
\includegraphics[width=1.\linewidth]{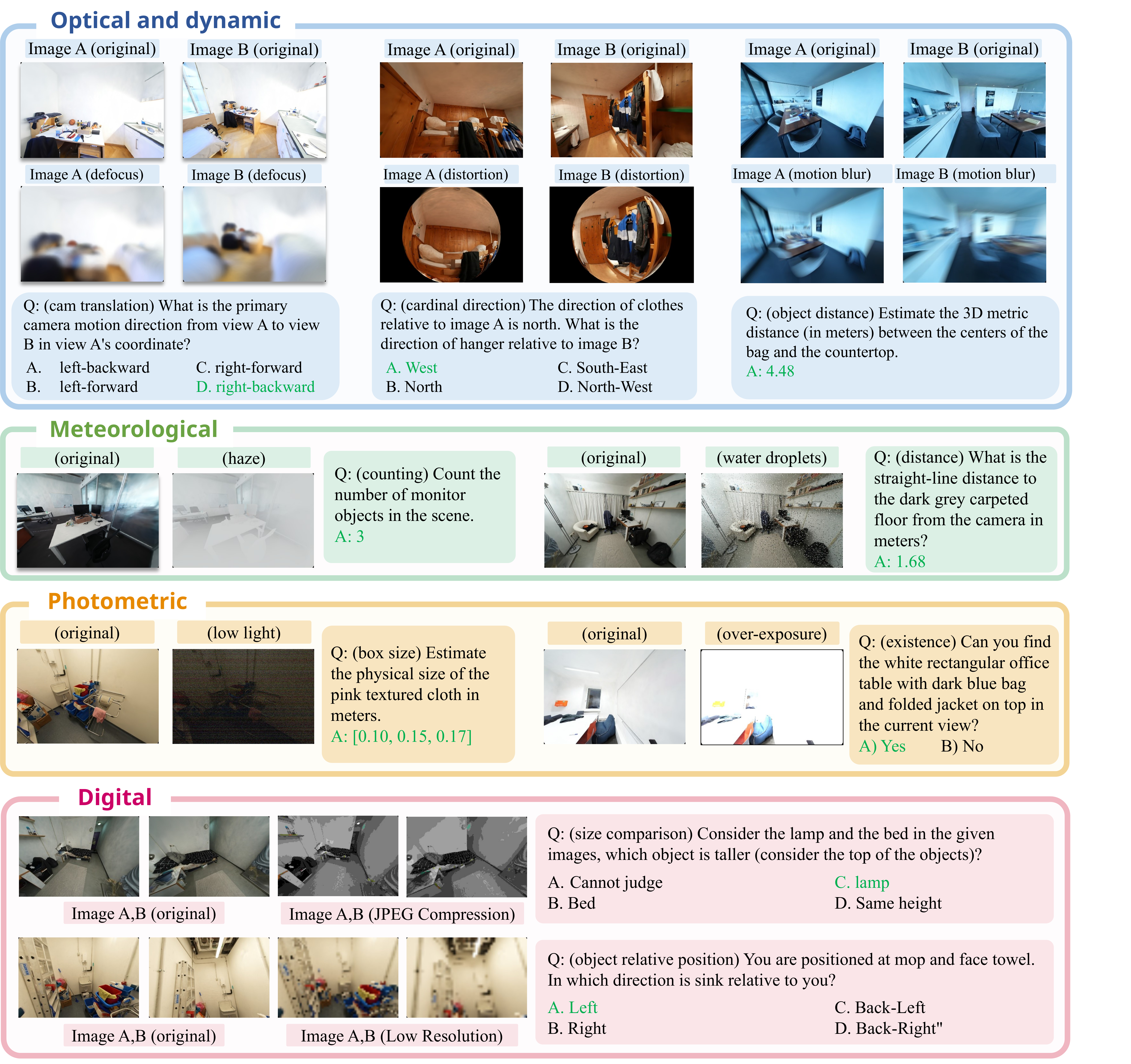}
\caption{Four degradation categories and their specific degradation QA Examples. We show both clean and degraded views with the corresponding spatial question which are simplified version without detailed object descriptions. For full QA examples please refer to Appendix~\ref{sec:appendix_examples}.}
\vspace{-3mm}
\label{fig:sample}
\end{figure*}

\begin{figure*}[t]
\centering
\includegraphics[width=1\linewidth]{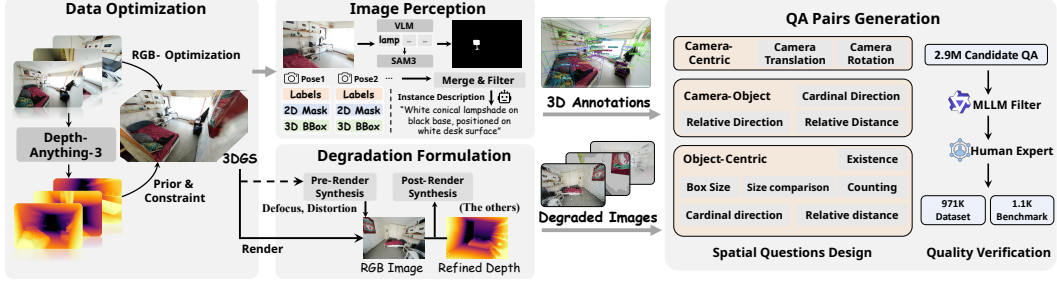}
\caption{The SpaceDG data engine. Input multi-view RGB images are first reconstructed into geometry-consistent 3DGS on which degradations are formulated. Pre-rendered degradations include defocus and distortion, while post-rendered degradations contain the others. Then SAM3 masks are lifted to 3D instances with generated descriptions. QA pairs are generated with structured templates and automatically-calculated answers, followed by a two-stage MLLM-plus-human verification.}
\label{fig:pipe}
\end{figure*}

\subsection{Data Engine}
\paragraph{3D Data Collection}
SpaceDG builds on the automatic 3D data curation pipeline of Holi-Spatial~\cite{gao2026holispatialevolvingvideostreams}, which converts raw video streams into geometry-consistent 3D semantic scenes. For each video, we first estimate depth and camera-pose priors with DepthAnything-v3~\cite{depthanything3} and COLMAP~\cite{schoenberger2016sfm} to optimize a geometrically constrained 3DGS representation. This gives us a renderable scene with calibrated camera poses and dense depth, which is critical for producing degradation variants without changing the underlying spatial ground truth. We then apply SAM3~\cite{carion2026sam} to key frames to obtain per-view semantic masks. The masks are lifted and associated across views using the reconstructed depth, camera poses, and bounding-box IoU, yielding object-level 3D instances with 3D bounding boxes, visible-frame lists, and the highest-confidence view for each instance.

\paragraph{QA Pairs Generation}
We initialize spatial QA pairs directly from the reconstructed 3D scene information. First, for each 3D instance we generate a short, view-invariant language description by asking VLM for its appearance in its highest-confidence SAM3 mask image. These descriptions allow questions to refer to natural objects without adding artificial markers like boxes or points on evaluated images\cite{deng2025internspatial}. Second, we sample valid single-view and two-view observations using pairwise image covisibility and minimum baseline constraints, so that each question is both visually answerable and geometrically non-trivial. Following MapAnything~\cite{keetha2026mapanything}, the covisibility score between two images is computed by reprojecting depth-supported 3D points from one calibrated view into the other and counting projections that pass a depth-based reprojection consistency check. Finally, we instantiate structured QA templates and compute answers from camera extrinsics, 3D box centers, object extents, and relative directions. This produces physically grounded answers for camera translation and rotation, object distance and direction, size comparison, and cross-view relational reasoning. Depending on the task, answers are represented as multiple-choice labels, binary decisions, or metric values. We provide QA examples in Figure~\ref{fig:sample}, and detailed generation rules in Appendix~\ref{sec:appendix_qa_generation}.

\paragraph{Degradation Synthesis}
Methods for compositing various degradations to RGB images have been thoroughly explored~\cite{9025509, wei2021physics, wang2024dofgs, steinrucken2017heartfelt, liao2024fisheyegslightweightextensiblegaussian}. To ensure physical realism and multi-view consistency, we further develop a physically grounded degradation pipeline that operates directly on 3DGS rendering process or linear light domain. We systematically inject 9 representative degradations across four categories: optical and dynamic degradations (defocus, distortion, motion blur), meteorological degradations (haze, water droplets), photometric degradations (low-light, over-exposure) and digital degradations (JPEG compression, low-resolution). As illustrated in Figure~\ref{fig:pipe}, all degradations are designed such that the underlying 3D spatial ground-truth remains invariant, ensuring accurate answers. We provide detailed formulations for each degradation process in Appendix~\ref{sec:appendix_degradation}.

\subsection{SpaceDG Dataset and SpaceDG-Bench}
\label{sec:spacedg_dataset_bench}

\begin{figure}[htbp]
    \centering

    \begin{minipage}[t]{0.48\textwidth}

        \vspace{10pt}
        \centering
        \renewcommand{\arraystretch}{1.1}
    
        \resizebox{\textwidth}{!}{
            \begin{tabular}{lcc}
                \toprule
                \textbf{Statistic} & \textbf{SpaceDG Dataset} & \textbf{SpaceDG-Bench} \\
                \midrule
                Unique questions & 971,090 & 1102 \\
                Unique images & 162,071 & 15525 \\
                Number of degradations & 9 & 9 \\
                Number of scenes & 584 & 320 \\
                Multi-degradations per question & \usym{2717} & \usym{2713} \\
                Single-view questions & 276,542 & 379 \\
                Multi-view questions & 694,548 & 723 \\
                Average image count per question & 1.72 & 1.56 \\
                Final VQA pairs & 971,090 & 9918 \\
                \bottomrule
            \end{tabular}
        }
        \vspace{8pt}
        \captionof{table}{Statistics of SpaceDG and SpaceDG-Bench. We report the number of unique questions, images, scenes, and degradation types, along with the breakdown of views.}\
        \label{tab:sta}
    \end{minipage}
    \hfill
    \begin{minipage}[t]{0.48\textwidth}
        \vspace{0pt}
        \centering
        \includegraphics[width=\textwidth]{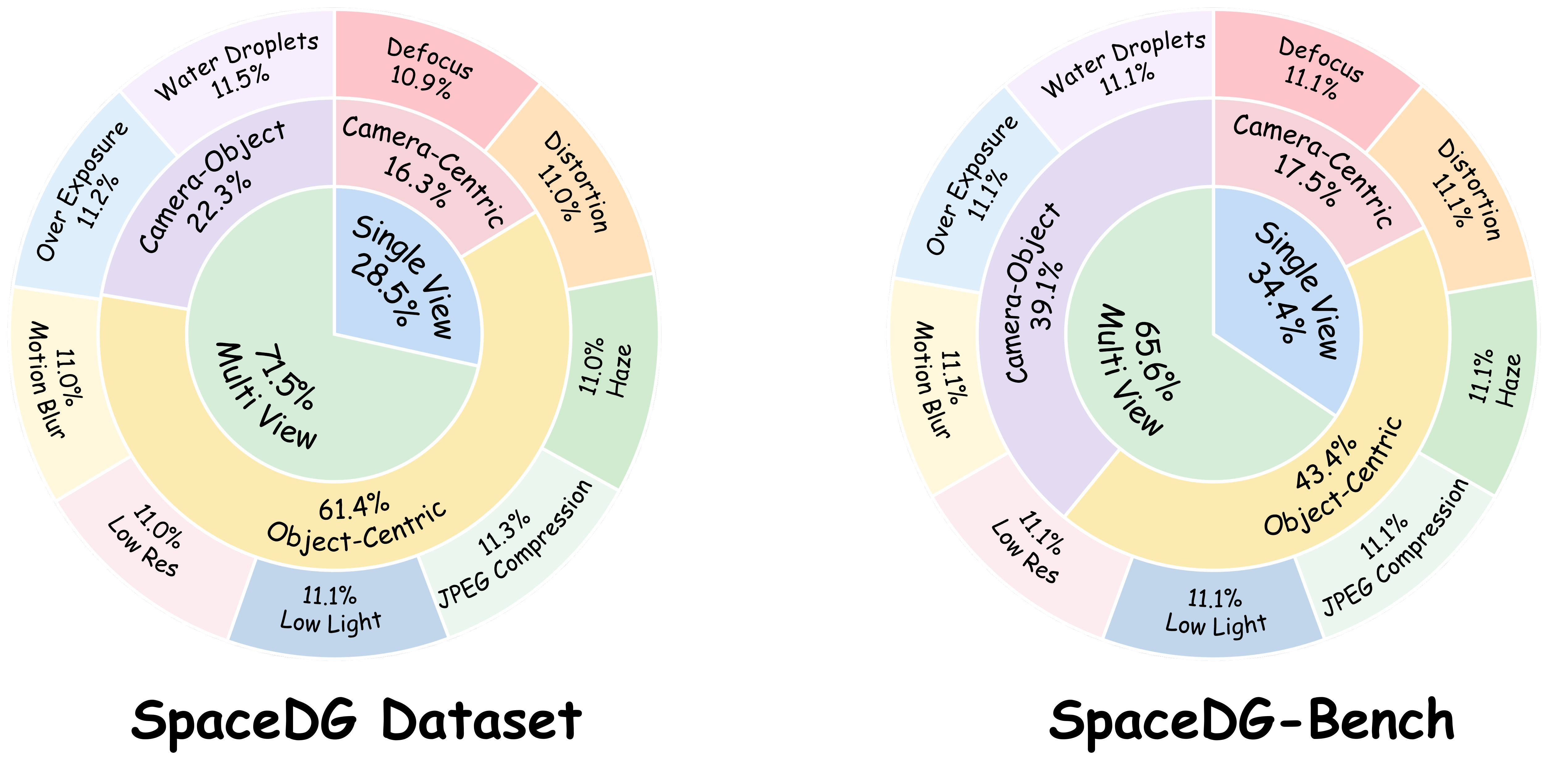}
        \captionof{figure}{Distribution of SpaceDG and SpaceDG-Bench. Inner-to-outer rings show the proportion of QA pairs by view configuration, spatial task group, and degradation type.}
        \label{fig:ring}
    \end{minipage}

\end{figure}

\paragraph{Statistics of SpaceDG}
Built upon the data engine, we construct SpaceDG dataset and SpaceDG-Bench. As shown in Table~\ref{tab:sta} and Figure~\ref{fig:ring}, SpaceDG dataset contains 971,090 QA instances, covering 584 real indoor scenes with physically synthesized degraded images. Each sample is organized as image observations and spatial questions with corresponding answers derived from geometry-consistent 3D annotations. We further curate SpaceDG-Bench from 320 representative scenes that are disjoint from the SpaceDG training set, resulting in 1,102 manually verified questions (723 multi-view and 379 single-view). For each benchmark item, we render one clean condition (original) and nine degraded conditions: defocus, distortion, haze, JPEG compression, low-light, low-resolution, motion blur, over-exposure, and water droplets. The benchmark is balanced at the image level, with 1,725 images per degraded condition, resulting in a benchmark with actual 9918 VQA pairs.

\paragraph{Spatial Questions Design}
To guarantee a comprehensive assessment of spatial intelligence, we systematically design 11 distinct question categories categorized into single-view and multi-view settings. These tasks evaluate three fundamental aspects: (1) Camera-centric, requiring models to estimate camera translation distance and relative rotations (e.g., yaw, pitch, and roll) between viewpoints; (2) Object-centric, encompassing object counting, object direction, distance estimation, and fine-grained 3D spatial extents ($[w, l, h]$); and (3) Camera-object Relational, which evaluates inter-object and camera-object spatial relations like cross-view direction and relative positioning.

\paragraph{Quality Verification}
To ensure data quality, we employ a two-stage filtering pipeline combining a VLM-based agent with human review. In the first stage, Qwen3-VL-32B serves as an automated judge to eliminate ambiguous questions — any question description that could plausibly refer to multiple objects or be incorrect is discarded. In the second stage, a human expert manually screens the remaining QA pairs through a dedicated interface, resulting in approximately 2,000 candidate pairs. Finally, two experts independently review the candidate set and remove QA pairs with ambiguous descriptions, incorrect answers, or ill-formed options.

\section{Experiments}
\label{sec:exp}

\begin{table*}[tbp]
\centering
\renewcommand{\arraystretch}{1.1}
\vspace{2mm}
\resizebox{\textwidth}{!}{
\begin{tabular}{l c|ccccccccc c}
\toprule
\multirow{2}{*}{\textbf{Models}} & \multirow{2}{*}{\begin{tabular}[c]{@{}c@{}}\textbf{Clean}\\\textbf{Image}\end{tabular}} & \multicolumn{9}{c}{\textbf{Degradation Types}} & \multirow{2}{*}{\textbf{Avg.}} \\ \cmidrule(lr){3-11}
& & Defocus & Distortion & Haze & JPEG-com. & Low-light & Low-res. & Motion-blur & Over-exp. & Water-droplets & Avg\\
\midrule
\multicolumn{12}{l}{\textit{\textbf{Base}}} \\
Human Level & 80.4 & 62.2 & 67.0 & 55.5 & 61.4 & 51.7 & 62.0 & 50.9 & 55.8 & 69.0    & 59.5 \\
Non-Image $_{\textit{Qwen3-VL-8B-Instruct}}$ & - & - & - & - & - & - & - & - & - & - & 33.5 \\
Non-Image $_{\textit{GPT-5.4}}$ & - & - & - & - & - & - & - & - & - & - & 35.1 \\
\midrule
\multicolumn{12}{l}{\textit{\textbf{Proprietary}}} \\
GPT-5.4 & 50.3 & 43.3 & 49.9 & 45.6 & 48.2 & 42.5 & 42.8 & 46.2 & 49.9 & 47.2 & 46.2 \\
Gemini-3.1-Flash-Lite & 56.9 & 44.2 & 55.7 & 46.1 & 52.8 & 43.1 & 48.0 & 46.5 & 52.3 & 50.4 & 48.8 \\
Gemini-3.1-Pro & 63.1 & 51.8 & 63.6 & 57.2 & 60.4 & 51.2 & 53.1 & 56.0 & 63.3 & 53.8 & 56.7 \\
Claude-Sonnet-4.6 & 52.4 & 44.8 & 52.5 & 39.5 & 49.8 & 37.9 & 40.9 & 43.9 & 49.0 & 39.7 & 44.2 \\
Grok-4.1-Fast & 39.7 & 34.8 & 37.9 & 35.5 & 37.3 & 33.1 & 35.1 & 34.8 & 36.3 & 35.5 & 35.6 \\
% Claude-Sonnet-4.6 & - & - & - & - & - & - & - & - & - & - & - \\
Qwen3.6-Plus & 58.3 & 40.9 & 54.8 & 46.0 & 49.9 & 37.8 & 43.6 & 45.2 & 50.1 & 45.4 & 46.0 \\
\midrule
\multicolumn{12}{l}{\textit{\textbf{Open-source general model}}} \\
InternVL3-8B & 42.5 & 37.0 & 41.9 & 37.7 & 41.5 & 38.2 & 41.4 & 39.5 & 42.4 & 40.9 & 40.1 \\
InternVL3-5-38B & 52.9 & 45.3 & 50.6 & 44.3 & 50.1 & 45.2 & 47.4 & 47.1 & 50.6 & 48.5 & 47.7 \\
InternVL3-5-8B & 46.7 & 39.2 & 45.2 & 38.6 & 44.1 & 36.3 & 40.8 & 41.2 & 44.6 & 43.0 & 41.4 \\
Llava-OneVision-Qwen2-7b-SI & 38.4 & 33.4 & 36.3 & 32.1 & 36.4 & 30.6 & 33.4 & 32.6 & 35.2 & 33.9 & 33.8 \\
Gemma-4-26B-A4B-it & 43.7 & 29.8 & 39.9 & 27.5 & 37.1 & 23.5 & 29.4 & 28.9 & 36.0 & 26.9 & 31.0 \\
Llama-4-Maverick & 41.1 & 31.6 & 39.4 & 34.3 & 37.4 & 30.5 & 33.8 & 29.9 & 35.8 & 35.7 & 34.3 \\
Kimi-VL-A3B-Instruct & 40.3 & 32.3 & 39.1 & 32.3 & 37.2 & 28.7 & 31.9 & 31.0 & 37.4 & 35.8 & 34.0 \\
Qwen3-VL-4B-Instruct & 48.5 & 37.2 & 44.1 & 37.5 & 43.1 & 34.8 & 38.5 & 38.1 & 43.7 & 40.4 & 39.7 \\
Qwen3-VL-8B-Instruct & 49.1 & 38.4 & 48.5 & 40.4 & 44.8 & 36.1 & 41.3 & 40.1 & 45.5 & 44.2 & 42.1 \\
Qwen3-VL-32B-Instruct & 55.0 & 43.8 & 53.9 & 41.9 & 49.2 & 36.8 & 42.3 & 43.6 & 49.4 & 45.9 & 45.2 \\
Qwen3.5-4B & 47.1 & 38.8 & 48.2 & 38.0 & 42.4 & 34.1 & 37.3 & 39.5 & 44.2 & 40.9 & 40.4 \\
Qwen3.5-9B & 49.3 & 38.4 & 46.7 & 38.3 & 45.4 & 37.3 & 41.0 & 40.6 & 46.8 & 41.2 & 41.7 \\
Qwen3.5-27B & 55.5 & 40.8 & 52.9 & 40.1 & 48.7 & 35.1 & 41.1 & 42.5 & 47.9 & 44.9 & 43.8 \\
Qwen3.6-35B-A3B & 53.9 & 40.3 & 50.9 & 38.1 & 47.3 & 33.4 & 39.8 & 41.4 & 47.5 & 42.6 & 42.4 \\
\midrule
\multicolumn{12}{l}{\textit{\textbf{Open-source spatial-intelligence model}}} \\
Cambrian-S-7B & 28.4 & 25.7 & 28.1 & 26.3 & 28.7 & 27.2 & 25.5 & 25.4 & 27.5 & 27.9 & 26.9 \\
VST-7B & 46.8 & 40.0 & 46.4 & 39.0 & 42.3 & 35.5 & 40.7 & 41.7 & 44.6 & 39.4 & 41.1 \\
SenseNova-SI-InternVL3-8B & 57.9 & 51.2 & 57.2 & 50.7 & 54.2 & 49.7 & 53.5 & 53.4 & 56.5 & 53.7 & 53.3 \\
\midrule
\multicolumn{12}{l}{\textit{\textbf{Open-source robotic brain}}} \\
ACE-Brain-0-8B & 50.2 & 43.1 & 48.9 & 45.1 & 46.7 & 41.2 & 42.7 & 43.8 & 47.5 & 46.6 & 45.1 \\
RynnBrain-8B & 51.7 & 45.6 & 50.5 & 41.3 & 47.1 & 41.0 & 43.7 & 44.9 & 47.5 & 45.3 & 45.2 \\
\midrule
\multicolumn{12}{l}{\textit{\textbf{Ours}}} \\
SpaceDG-SFT$_{\textit{InternVL-3.5-8B}}$ & 70.9 & 64.6 & 67.6 & 62.2 & 67.5 & \textbf{61.1} & \textbf{64.3} & 64.4 & 68.5 & 66.7 & 65.2 \\
SpaceDG-SFT$_{\textit{Qwen3-VL-8B-Instruct}}$ & \textbf{73.2} & \textbf{65.6} & \textbf{69.2} & \textbf{64.8} & \textbf{68.6} & 59.2 & 63.9 & \textbf{66.1} & \textbf{70.3} & \textbf{67.5} & \textbf{66.1} \\
\bottomrule
\end{tabular}
}
\caption{Quantitative comparison of models on SpaceDG-Bench. We evaluate proprietary, open-source general, spatial-intelligence, and robotic-brain models under the clean condition and nine visual degradations, together with human-level and non-image baselines.}
\label{tab:main}
\end{table*}

\subsection{Evaluation Setup}
\paragraph{Baselines}
We systematically evaluate 25 models on SpaceDG-Bench, including proprietary models like GPT-5.4~\cite{openai2026gpt54}, Gemini-3.1-Pro~\cite{google2025gemini31pro}, Gemini-3.1-Flash-Lite~\cite{google2025gemini31flash}, Claude-Sonnet-4.6~\cite{anthropic2026claudesonnet46}, open-source general models like Qwen3.5~\cite{qwen3.5}, InternVL3.5~\cite{wang2025internvl35advancingopensourcemultimodal}, Kimi-VL~\cite{kimiteam2025kimivltechnicalreport}, LLaVA-OneVision-1.5~\cite{LLaVA-OneVision-1.5} and so on. We also evaluate domain-specific models like spatial-intelligence models~\cite{sensenova-si, vst, yang2025cambrians} and robotic brains~\cite{gong2026ace, damo2026rynnbrain}. For each model, we evaluate it on both clean images and 9 visual degradations using EASI~\cite{easi2025} and VLMEvalKit~\cite{duan2024vlmevalkit} under zero-shot settings. We also include two baselines: human-level assessment and non-image on GPT-5.4 and Qwen3-VL-8B-Instruct.

\paragraph{Evaluation Metrics}
To rigorously evaluate these heterogeneous answers, we adopt two metrics tailored to the output formats. For multiple-choice questions (MCQ) and binary-decision questions, we report Accuracy (Acc). For numerical answer (NA) questions requiring exact metric scalars (e.g., distances and sizes), we employ Mean Relative Accuracy (MRA)~\cite{yang2024think} with confidence thresholds $\Theta = \{0.50, 0.55, \dots, 0.95\}$. For list-type numerical answer (e.g., size estimation), we require the model to output answers in ascending order and calculate the metric by weighting each number.

\subsection{Evaluation Results on SpaceDG-Bench}
\begin{wrapfigure}{t}{0.4\textwidth}
    \small
    \vspace{-3mm}
    \centering
    \includegraphics[width=\linewidth]{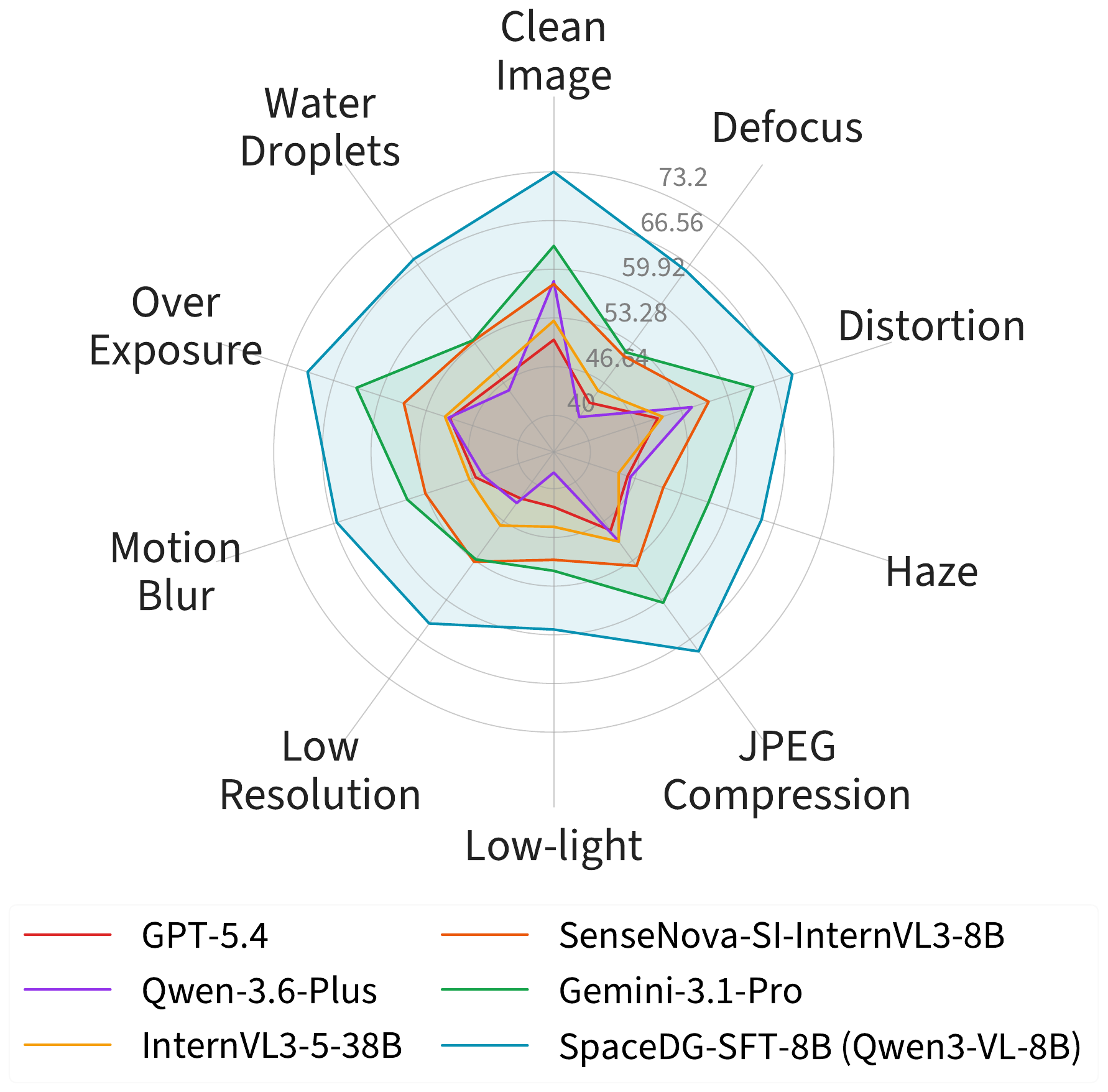}
    \caption{
        Per-degradation performance of representative models.
    }
    \label{fig:degradation_radar}
    % \vspace{-0.1mm}
\end{wrapfigure}

\paragraph{Visual degradation consistently impairs spatial reasoning.} 
As shown in Table~\ref{tab:main}, all evaluated MLLMs achieve lower performance under degraded inputs than under clean images, demonstrating that spatial intelligence remains highly sensitive to realistic visual corruptions. For instance, Gemini-3.1-Pro performs best among tested proprietary models, achieving 63.1\% on clean images but decreases to 56.7\% on degraded images. Qwen3.6-Plus exhibits strong performance on clean images but suffers a severe performance decrease under degraded conditions, especially on defocus and low light, as shown in Figure~\ref{fig:degradation_radar}. Open-source models exhibit a similar trend: for example, InternVL3.5-38B drops from 52.9\% on clean images to 47.7\% under degraded inputs, revealing a substantial performance gap between ideal and degraded visual conditions. Additionally, we report the performance of GPT-5.4 and Qwen-3-VL-8B-Instruct when provided with no input image, and both models perform significantly worse than their degraded-image counterparts, approaching random-guess level. This result confirms that our benchmark contains few exploitable language shortcuts, and that degraded images still retain rich visual information necessary for correct answers.

\paragraph{Humans struggle with extreme visual degradation.}
To establish a human reference baseline, we evaluate human performance on a 900-question subset of SpaceDG-Bench. As shown in Table~\ref{tab:main}, humans achieve 80.4\% accuracy on clean images, substantially outperforming all evaluated MLLMs. However, their performance drops by 20.9\% under degraded conditions, indicating that severe visual corruptions can significantly impair fine-grained spatial judgment. These results suggest that spatial reasoning under degradation is challenging not only for current MLLMs but also for human observers, underscoring the need for degradation-aware training and evaluation protocols. Details of the human study are provided in Appendix~\ref{sec:appendix_human}.

\paragraph{Degradation-aware SFT effectively improves the performance of MLLMs.}
We utilize the constructed SpaceDG to conduct supervised fine-tuning on Qwen-3-VL-8B-Instruct and InternVL-3.5-8B for 1 epoch with a batch size of 2048 using 8$\times$H200 GPUs. Our SpaceDG-SFT-Qwen3 achieves substantial improvements over its base model across both clean and degraded conditions, rising from 49.1\% to 73.2\% on clean images and from 42.1\% to 66.1\% on degraded inputs. Notably, under degraded conditions, SpaceDG-SFT-Qwen3 surpasses the human reference performance of 59.5\% by 6.6 percentage points. These results provide two key insights into degradation-aware spatial intelligence. First, supervised fine-tuning with degradation-augmented data substantially improves the spatial reasoning capability and robustness of MLLMs, suggesting that degradation-aware training is a practical path toward robust real-world spatial intelligence. Second, the gap between human and model performance under degraded conditions indicates that severe visual corruptions can also limit human spatial judgment, while models trained on large-scale degradation-aware data can learn to better exploit visual cues in challenging observations.

\begin{table*}[htbp] 
\centering 
\renewcommand{\arraystretch}{1.1} 
\vspace{2mm} 
\resizebox{\textwidth}{!}{ 
\begin{tabular}{l ccccccccc c} 
\toprule 
\multirow{2}{*}{\textbf{Models}} & \multicolumn{9}{c}{\textbf{Degradation Types ($\Delta$)}} & \multirow{2}{*}{\textbf{Avg. $\Delta$}} \\ \cmidrule(lr){2-10} 
& Defocus & Distortion & Haze & JPEG-com. & Low-light & Low-res. & Motion-blur & Over-exp. & Water-droplets & \\ 
\midrule 
Llava-OneVision-Qwen2-7b-SI & +0.4 & -0.1 & +1.5 & +0.5 & +1.6 & -0.6 & +0.4 & +0.5 & +1.1 & +0.6 \\
Qwen3-VL-4B-Instruct & +1.7 & +1.3 & +1.5 & +0.9 & +0.4 & +1.3 & +0.8 & +2.5 & +2.5 & +1.4 \\ 
Qwen3-VL-8B-Instruct & +2.4 & +0.6 & -1.0 & +1.2 & +1.1 & -0.9 & +0.4 & +2.5 & +0.6 & +0.8 \\ 
Qwen3-VL-32B-Instruct & +1.4 & -0.1 & +0.8 & -0.3 & +1.5 & +1.4 & +1.2 & +0.6 & -0.8 & +0.6 \\ 
Qwen3.5-9B & +0.6 & +2.0 & +0.7 & +1.0 & +2.3 & -1.6 & +1.1 & +0.3 & +0.2 & +0.7 \\ 
RynnBrain-8B & -1.8 & +0.1 & -0.1 & +0.5 & +0.0 & -1.5 & -0.5 & +1.1 & -1.3 & -0.4 \\ 
ACE-Brain-0-8B & -1.2 & +0.3 & -0.6 & -0.3 & +1.6 & +0.1 & +0.2 & -0.0 & -0.1 & -0.0 \\ 
VST-7B & -0.2 & -1.8 & -0.9 & -0.1 & +0.6 & +0.1 & -0.8 & -1.3 & -0.1 & -0.5 \\ 
SenseNova-SI-InternVL3-8B & -1.3 & +0.4 & -0.7 & +0.2 & -0.1 & -0.9 & -0.4 & +0.1 & +0.3 & -0.3 \\ 
SpaceDG-SFT-Qwen3-VL-8B & -1.1 & -0.9 & +2.5 & -0.6 & +0.4 & +0.9 & +0.6 & -1.3 & -0.5 & -0.0 \\
\bottomrule 
\end{tabular} 
}
\caption{Performance changes when degradation type and severity are explicitly provided by prompts. We use the degradation prompt template in Appendix~\ref{sec:appendix_dgprompt}.} 
\label{tab:delta} 
\end{table*}

\paragraph{Spatial fine-tuning enhances the visual robustness of MLLMs, while reducing degradation comprehension capability.}
As shown in Table~\ref{tab:main}, spatially fine-tuned and robotic brain models exhibit a smaller performance drop when transitioning from clean to degraded inputs. On average, these models decline by 5.5\%, compared to 7.6\% for general models, indicating stronger inherent robustness to visual corruptions. However, Table~\ref{tab:delta} reveals an opposing trend in degradation comprehension capability: when the degradation type and severity are explicitly provided in the prompt, general-purpose models consistently benefit, achieving notable performance gains across all degradation categories. In contrast, spatially fine-tuned models show little to no improvement, with some even exhibiting a slight performance decrease. This suggests that spatial fine-tuning encourages models to develop degradation-agnostic visual representations, trading away sensitivity to image quality cues in favor of task-level robustness.

\begin{figure*}[ht]
\includegraphics[width=1\linewidth]{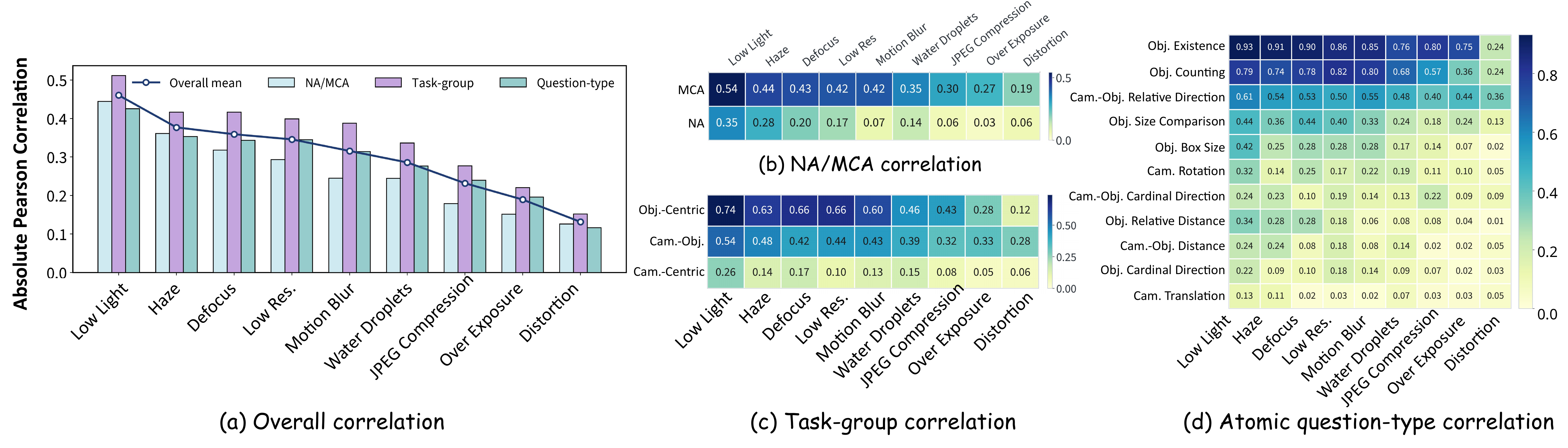}
\caption{Degradation-wise correlation analysis. Sensitivity of spatial intelligence is measured by the absolute point-biserial correlation \(|r|\). (a) Overall \(|r|\) per degradation. (b) Breakdown by answer format. (c) Breakdown by task group. (d) Per-atomic-question correlation.}
% \vspace{-0.3cm}
\label{fig:cor}
\end{figure*}

\subsection{Degradation-wise Correlation Analysis}
\paragraph{Metric Design}
To quantify the sensitivity of spatial reasoning performance to various image degradations, we adopt the absolute point-biserial Pearson correlation coefficient, $|r|$. For each analysis slice (e.g., answer format, task group, or specific question type) and degradation type, we construct paired observations across all evaluated models. Let the binary indicator $\mathbf{x} \in \{0, 1\}$ denote whether a score originates from a clean ($x=0$) or degraded ($x=1$) condition, and $\mathbf{y}$ represent the concatenated score vector across all $M$ models from Table~\ref{tab:main}:
\begin{equation}
    \mathbf{x}=[0,\dots,0,1,\dots,1], \quad
    \mathbf{y}=[s^{(1)}_{\text{ori}},\dots,s^{(M)}_{\text{ori}},s^{(1)}_{\text{deg}},\dots,s^{(M)}_{\text{deg}}].
\end{equation}
The correlation is computed as:
\begin{equation}
    r = \mathrm{corr}(\mathbf{x}, \mathbf{y}) = \frac{\sum_{i}(x_i - \bar{x})(y_i - \bar{y})}{\sqrt{\sum_{i}(x_i-\bar{x})^2}\sqrt{\sum_{i}(y_i-\bar{y})^2}}.
\end{equation}
We report $|r|$ to reflect the magnitude of the degradation effect; a larger $|r|$ indicates a more significant score shift between clean and degraded conditions. The results are shown in Figure~\ref{fig:cor}.

\paragraph{Analysis}
Across all subfigures in Figure~\ref{fig:cor} we can observe that low-light and haze consistently induce the most pronounced performance drops across models, whereas over-exposure and distortion have comparatively weaker effects. Figure~\ref{fig:cor} (b) further shows that Multiple-Choice Answer (MCA) questions exhibit higher degradation correlation than Numerical Answer (NA) questions. The task-group analysis in Figure~\ref{fig:cor} (c) indicates that object-centric tasks are the most sensitive to visual degradations, while camera-centric tasks remain relatively robust, suggesting that localized object grounding is more severely disrupted than global scene-level perception. At the atomic question-type level, fine-grained semantic perception tasks, such as existence estimation and object counting, exhibit the highest correlation, whereas tasks requiring global understanding, such as camera translation, show the lowest correlation. These findings suggest that \textbf{visual degradations primarily impair MLLMs' fine-grained semantic perception}, thereby disproportionately affecting tasks that require detailed visual grounding.

\begin{figure*}[t]
\centering
\includegraphics[width=0.95\linewidth]{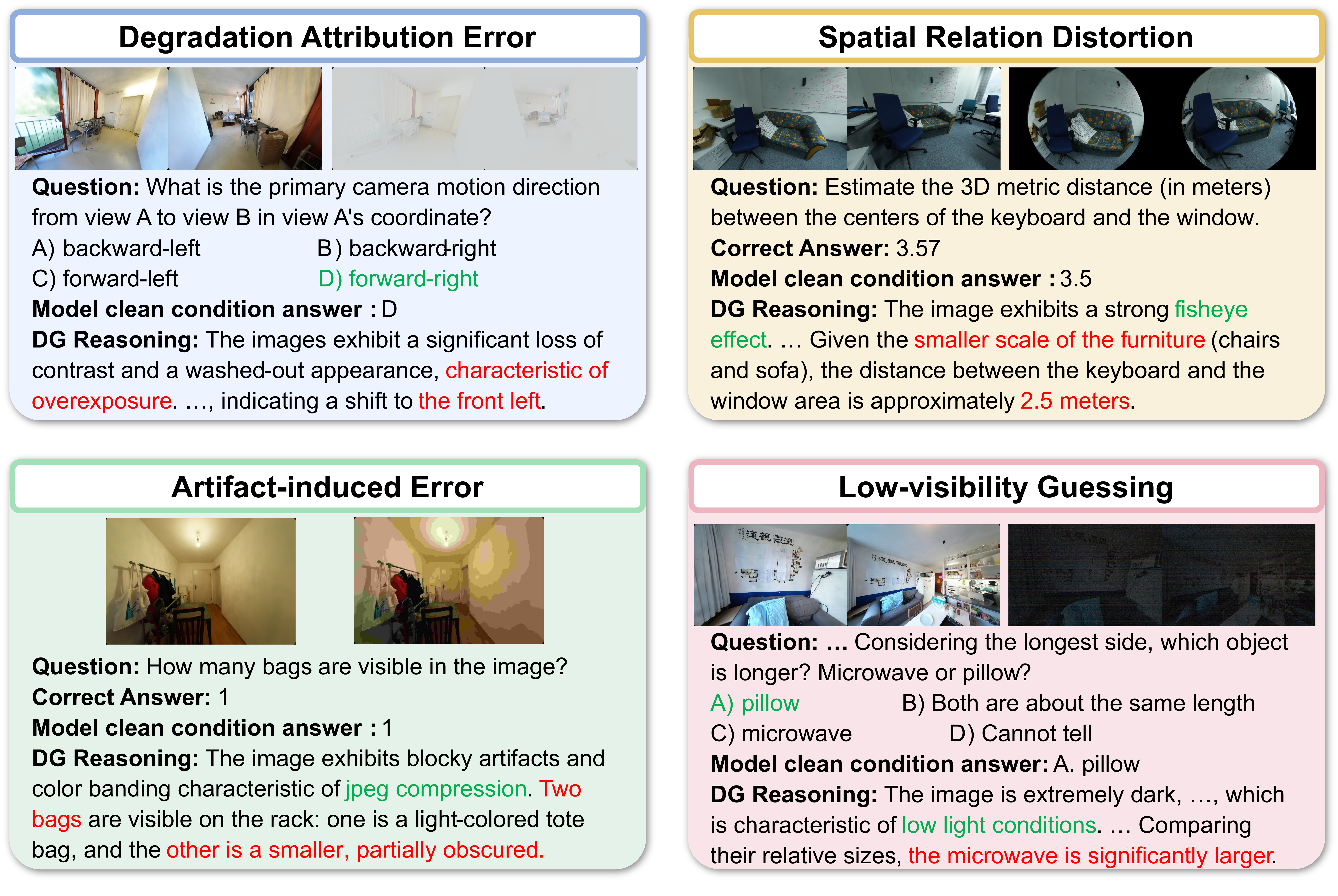}
\caption{The four identified errors caused by visual degradations. We provide the correct answer, model answers under clean condition and reasoning processes under degraded condition.}
% \vspace{-3mm}
\label{fig:case}
\end{figure*}

\section{Degradation-guided Spatial Reasoning}
\subsection{Two-stage Chain-of-Thought Reasoning}

To investigate the impact of Chain-of-Thought (CoT)~\cite{wei2023chainofthoughtpromptingelicitsreasoning} for visual degradations, we design a structured two-stage CoT prompt. The MLLM is first required to explicitly output the degradation type from the provided list with a short description, followed by a classic prompt for reasoning. We conduct the experiment on Gemini-3.1-Flash-Lite and provide the detailed prompt in Appendix~\ref{sec:appendix_cot}. As shown in Table~\ref{tab:cot}, the model with CoT suffers from a 1.8\% decrease, demonstrating the harmness of CoT for degradation-guided reasoning.

\begin{table}[htbp]
\centering
\vspace{1mm}
\begin{tabular}{ccc}
\hline
Model & Method & Performance \\
\hline
\multirow{2}{*}{Gemini-3.1-Flash-Lite} 
    & w/o CoT  & 48.8 \\
    & with CoT & 47.0 \\
\hline
\end{tabular}
\vspace{1mm}
\caption{CoT performance of Gemini-3.1-Flash-Lite on SpaceDG-Bench.}
\label{tab:cot}
\vspace{-2mm}
\end{table}

\subsection{Error Analysis}
As shown in Figure~\ref{fig:case}, we further systematically examine the reasoning processes of Gemini-3.1-Flash-Lite and categorize degradation-induced errors into four types: 
(1) \textbf{degradation attribution errors}, where the model misidentifies the underlying corruption type (e.g., mistaking haze for over-exposure), leading to an incorrect reasoning premise from the first step; 
(2) \textbf{spatial relation distortion}, where degraded visual cues bias orientation and relative-position judgments, causing systematic errors in directional and relational reasoning; 
(3) \textbf{artifact-induced errors}, where compression artifacts, blur, and low-resolution textures introduce spurious patterns that mislead object counting and numeric/metric estimation; and 
(4) \textbf{low-visibility guessing}, where the model acknowledges poor observability but still produces overconfident answers instead of performing calibration or giving a conservative answer. While these errors reflect different shortcomings of the model, a sample may contain multiple errors.

\section{Conclusion}
In this work, we study spatial intelligence of MLLMs under realistic visual degradations, a setting that is critical for real-world embodied and autonomous systems but largely overlooked by existing spatial reasoning benchmarks. We introduce SpaceDG, a large-scale dataset constructed with a physically grounded degradation synthesis engine built on 3DGS, and SpaceDG-Bench, a human-verified benchmark covering diverse spatial reasoning categories and nine representative degradation types. Through a comprehensive evaluation of 25 proprietary, open-source, spatially fine-tuned, and robotic-brain models, we show that visual degradations consistently impair spatial reasoning, revealing a substantial robustness gap between clean and imperfect visual conditions. Our analysis further shows that fine-grained object-level perception is particularly vulnerable to degradations, while certain global geometric reasoning tasks remain relatively more robust. Finally, we demonstrate that supervised fine-tuning on SpaceDG substantially improves both clean and degraded performance, suggesting that degradation-aware training is a promising direction for building robust spatially intelligent MLLMs. We hope SpaceDG and SpaceDG-Bench will facilitate future research on spatial reasoning beyond idealized visual inputs and encourage the development of models that can reason reliably under imperfect real-world observations.
% \newpage

{
\small
\bibliographystyle{unsrt}
\bibliography{ref}
}
\newpage

%%%%%%%%%%%%%%%%%%%%%%%%%%%%%%%%%%%%%%%%%%%%%%%%%%%%%%%%%%%%

\appendix

\section{Additional Experiments}

\subsection{Data Validation}
\paragraph{Degradation-aware SFT enhances robust spatial reasoning.}
To isolate the effect of degradation augmentation from the general benefit of SFT, we compare degradation-augmented SFT against a clean-image SFT with the same training settings. As shown in Table~\ref{tab:transferability}, both achieve nearly identical performance on clean inputs (73.2\% vs. 73.1\%), indicating comparable spatial understanding ability under pristine conditions. However, under degraded inputs, degradation-augmented SFT improves the average performance from 64.6\% to 66.1\%, demonstrating that robustness gains from degradations. Further held-out experiments show that models trained without specific degradation categories still generalize well to unseen corruptions, substantially outperforming the no-SFT baseline. These results suggest that degradation-aware SFT encourages degradation-agnostic spatial reasoning strategies and is essential for robust real-world spatial intelligence.

\begin{table}[ht]
\centering
\resizebox{\linewidth}{!}{%
\begin{tabular}{lc|ccc|cc|cc|cc|c}
\toprule
\multirow{2}{*}{\textbf{Training Method}} & \multirow{2}{*}{\begin{tabular}[c]{@{}c@{}}\textbf{Clean}\\\textbf{Image}\end{tabular}} & \multicolumn{9}{c|}{\textbf{Degradation Types}} & \multirow{2}{*}{\textbf{Avg.}} \\
\cmidrule(lr){3-11}
& 
& Defocus 
& Distortion 
& Motion-blur 
& Haze 
& Water-droplets
& Low-light 
& Over-exp.
& JPEG-com. 
& Low-res. &  \\
\midrule
No-SFT
& 49.1
& 38.4 & 48.5 & 40.4 & 44.8 & 36.1 & 41.3 & 40.1 & 45.5 & 44.2 & 42.1 \\

Full-SFT with degradations 
& 73.2 
& 65.6 & 69.2 & 66.1 & 64.8 & 67.5 & 59.2 & 70.3 & 68.6 & 63.9 & 66.1 \\

Full-SFT with clean images 
& 73.1 
& 63.2 & 68.3 & 64.3 & 65.4 & 65.9 & 58.3 & 67.1 & 67.0 & 64.2 & 64.8 \\
\midrule

\textbf{Held-out Degradation Types} \\

w/o Optical \& Dynamic 
& Defocus, Distortion, Motion Blur
& 64.0 & 69.9 & 65.2 \\

w/o Meteorological
& Haze, Water Droplets
&  &  &  & 64.3 & 66.02 \\

w/o Photometric
& Low Light, Over Exposure
& & & & & & 57.8 & 66.1 \\

w/o Digital
& JPEG Compression, Low Resolution
& & & &  & &  & & 67.4 & 63.8 \\

\bottomrule
\end{tabular}%
}
\vspace{2mm}
\caption{Ablation study of SpaceDG SFT. Using Qwen3-VL-8B-Instruct as the base model, we compare the no-SFT baseline, full SFT with and without degraded images, and held-out variants that exclude each degradation category at a time.}
\label{tab:transferability}
\vspace{-4mm}
\end{table}

\subsection{Real-world Inspired Mixture of Degradations}
Real-world images rarely suffer from a single isolated corruption; instead, multiple degradations often co-occur due to complex acquisition conditions. To evaluate robustness under such compound effects, we extend SpaceDG from single-degradation evaluation to a real-world inspired mixture protocol. We design six mixture recipes that reflect representative capture scenarios: night capture, hazy long-range observation, wet-lens motion, backlit dynamic scenes, motion-defocus, and compressed portrait sharing. In each recipe, we pick one primary degradation applied the same degradation settings with SpaceDG-Bench, while all auxiliary degradations are applied at easier severity level. The primary degradation for each recipe is highlighted in Table~\ref{tab:mix_degradation_results}.

As shown in Table~\ref{tab:mix_degradation_results}, compound degradations substantially challenge the base model, whose performance drops to an average accuracy of 37.0 across the six mixture settings. In contrast, SpaceDG-SFT-Qwen3-VL-8B-Instruct achieves consistently higher performance in all scenarios, achieving an average accuracy of 62.4. The gains are especially pronounced under hazy long-range observation, compressed portrait sharing, and motion-defocus, suggesting that SpaceDG training improves not only robustness to individual corruptions but also generalization to realistic combinations of multiple visual degradations. These results indicate that spatial reasoning remain sensitive to compound image degradation, while degradation-aware training can substantially enhance their reliability in real-world visual conditions.

\begin{table}[htbp]
\centering
\small
\resizebox{\textwidth}{!}{ 
\begin{tabular}{lcccccc}
\toprule
\multirow{2}{*}{Models}
& Night Capture
& Hazy Long-range
& Wet-lens Motion
& Backlit  Dynamics
& Motion-Defocus
& Compressed Portrait \\
\cmidrule(lr){2-7}
& \shortstack{\textbf{LL}+MB+LR}
& \shortstack{\textbf{HZ}+LR+MB}
& \shortstack{\textbf{WD}+MB+LL}
& \shortstack{\textbf{OE}+MB}
& \shortstack{\textbf{MB}+DF+LR}
& \shortstack{\textbf{DF}+JPEG} \\
\midrule
Qwen3-VL-8B-Instruct & 31.5 & 36.2 & 32.5 & 44.7 & 39.8 & 37.4 \\
SpaceDG-SFT-Qwen3-VL-8B-Instruct & 55.8 & 65.1 & 55.6 & 68.3 & 65.7 & 63.8 \\
\bottomrule
\end{tabular}
}
\vspace{2mm}
\caption{Performance under six mixed-degradation settings. Each column corresponds to one real-world inspired mixture recipe. The degradation in bold is the primary corruption. ``LL'', ``MB'', ``LR'', ``HZ'', ``WD'', ``OE'', ``DF'', and ``JPEG'' denote low-light, motion blur, low resolution, haze, water droplets, over-exposure, defocus, and JPEG compression, respectively.}
\label{tab:mix_degradation_results}
\end{table}

\subsection{Will supervised fine-tuning on degraded dataset affect the performance of general benchmarks?}

To investigate whether fine-tuning on a degraded dataset adversely affects spatial capabilities on clean images, we evaluate SpaceDG-SFT-8B on two general benchmarks: MMSI-Bench and MindCube. As shown in Table~\ref{tab:spatial_comparison}, SpaceDG-SFT-8B achieves scores of 30.0 and 37.0, respectively. These results outperform several baselines of comparable scale, including Qwen3-VL-8B and SpaceI-SFT-7B, and remain competitive with stronger models such as Intern3-VL-8B and VST-SFT-7B. These findings suggest that fine-tuning on the degraded dataset does not significantly compromise the model's general capabilities.

\begin{table}[ht]
\centering
\small
\caption{Extra comparison on general benchmarks MMSI-Bench and MindCube.}
\vspace{2mm}
\label{tab:spatial_comparison}
\setlength{\tabcolsep}{12pt}
\renewcommand{\arraystretch}{1.08}
\begin{tabular}{lcc}
\toprule
\textbf{Model} & \textbf{MMSI-Bench} & \textbf{MindCube} \\
\midrule
VST-SFT-3B     & 30.2 & 35.9 \\
Cambrian-S-3B & 25.2 & 32.5 \\
VST-SFT-7B       & 32.0 & 39.7 \\
Cambrian-S-7B & 25.8 & 39.6 \\
SpaceI-SFT-7B  & 27.4 & 37.9 \\
Intern3-VL-8B & 28.0 & 41.5 \\
Spatial-MLLM & 27.0 & 32.1 \\
Qwen3-VL-8B  & 31.1 & 29.4 \\
\midrule
\addlinespace[2pt]

SpaceDG-SFT-Qwen3-VL-8B-Instruct          & 31.3 & 37.0 \\
\bottomrule
\end{tabular}
\end{table}

\subsection{Human-level Assessment}
\label{sec:appendix_human}
We conduct human-level assessment on SpaceDG-Bench-900, a 900-question subset of SpaceDG-Bench. To validate the reliability of this subset, we compare the performance of 13 representative models on both the full SpaceDG-Bench (9,918 samples) and this 900-sample subset (100 questions per degradation). As detailed in Table~\ref{tab:spacedg_900}, the models exhibit highly consistent performance across both evaluation sets, yielding an average absolute performance difference of merely 0.83\%. Specifically, the maximum performance gap is only 1.8\% (Gemini-3.1-Pro), and the minimum discrepancy is as low as 0.3\% (GPT-5.4). This negligible variance empirically demonstrates that SpaceDG-Bench-900 preserves the data distribution and task difficulty of the full evaluation benchmark. 

During the evaluation, we divide human annotators into two groups and assess their performance separately under clean and degraded conditions. On clean images, humans achieve an overall accuracy of 80.4\%, with 90.2\% accuracy on multiple-choice answer (MCA) questions, 61.2\% on numerical-answer (NA) questions, and 48.1\% on list-type NA questions. This breakdown suggests that humans perform well on general spatial reasoning questions but struggle with tasks requiring precise metric estimation. Human performance under each degradation type is summarized in Table~\ref{tab:main}.

\begin{table}[htbp]
    \centering
    \small
    \caption{Model Performance on SpaceDG-Bench and SpaceDG-Bench-900. We present this result to demonstrate the reliability of human benchmarks.}
    \label{tab:spacedg_900}
    \begin{tabular}{lcc} % l=left, c=center, c=center
        \toprule
        \textbf{Model} & \textbf{SpaceDG-Bench} & \textbf{SpaceDG-Bench-900} \\
        \midrule
        Number of samples & 9918 & 900 \\
        \midrule
        GPT-5.4                & 46.2 & 46.5 \\
        Claude-Sonnet-4.6     & 44.2 & 43.4 \\
        Gemini-3.1-Pro        & 56.7 & 54.9 \\
        Gemini-3.1-Flash-Lite  & 48.8 & 48.3 \\
        Qwen3-VL-4B-Instruct & 39.7 & 38.7 \\
        Qwen3-VL-8B-Instruct & 42.1 & 42.9 \\
        InternVL3-5-8B & 41.4 & 42.2 \\
        InternVL3-5-38B & 47.7 & 48.8 \\
        Qwen3.5-4B & 40.4 & 39.7 \\
        Qwen3.5-9B & 41.7 & 42.5 \\
        Qwen3.6-35B-A3B & 42.4 & 41.7 \\
        SenseNova-SI-InternVL3-8B & 53.3 & 52.8 \\
        ACE-Brain-0-8B & 45.1 & 44.1 \\
        \bottomrule
    \end{tabular}
\end{table}

\subsection{Degradation Prompt Template}
\label{sec:appendix_dgprompt}
\begin{figure*}[h]
\centering
\includegraphics[width=1\linewidth]{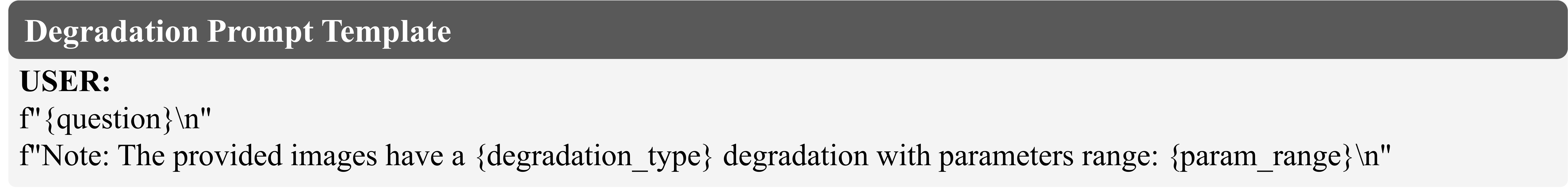}
\caption{Two-stage prompt template for degradation-guided spatial reasoning.}
% \vspace{-0.3cm}
\label{fig:dg_prompt}
\end{figure*}

We use the prompt template shown in Figure~\ref{fig:dg_prompt} to explicitly provide the MLLM with degradation-aware information during evaluation. Specifically, for each degraded input, the prompt augments the original spatial question with the degradation type and its corresponding severity, represented by the rendering parameter range. The parameter ranges used to generate SpaceDG-Bench are summarized in Table~\ref{tab:param_ranges}, and correspond to the degradation formulations introduced in Section~\ref{sec:appendix_degradation}, covering optical and dynamic, meteorological, photometric, and digital degradation processes. Reporting these settings makes the degradation-guided evaluation protocol and benchmark rendering configuration explicit and reproducible.

\begin{table}[t]
\centering
\resizebox{\linewidth}{!}{%
\begin{tabular}{l l}
\toprule
Degradation & Parameter ranges \\
\midrule
defocus &
aperture $\in [10.0,\,15.0]$, depth $\in [1.0,\,8.0]$ \\
distortion &
k1 $\in [-0.24,\,-0.23]$, k2 $\in [0.0001,\,0.0003]$, k3 $\in [0.0001,\,0.0002]$, k4 $\in [0.0000,\,0.0001]$, max\_theta $= 1.5$ \\
haze &
density $\in [3.5,\,6.0]$ \\
jpeg\_compression &
quality $\in [2,\,5]$ \\
low\_light &
exposure $\in [0.003,\,0.005]$ \\
low\_res &
scale $\in [0.02,\,0.05]$ \\
motion\_blur &
trans $\in [0.2,\,0.35]$, rot $\in [0.06,\,0.12]$, sub\_steps $\in [80,\,80]$ \\
over\_exposure &
exposure $\in [7.0,\,10.0]$ \\
water\_droplets &
scale $\in [2.5,\,4.0]$, radius $\in [0.25,\,0.75]$, strength $\in [0.3,\,0.5]$, blur\_sigma $\in [2.0,\,2.5]$, blur\_kernel $= 9$ \\
\bottomrule
\end{tabular}%
}
\vspace{1mm}
\caption{Parameter ranges for each degradation setting in SpaceDG-Bench.}
\label{tab:param_ranges}
\end{table}

\subsection{Degradation-guided Chain-of-Thought}
\label{sec:appendix_cot}
\begin{figure*}[h]
\centering
\includegraphics[width=1\linewidth]{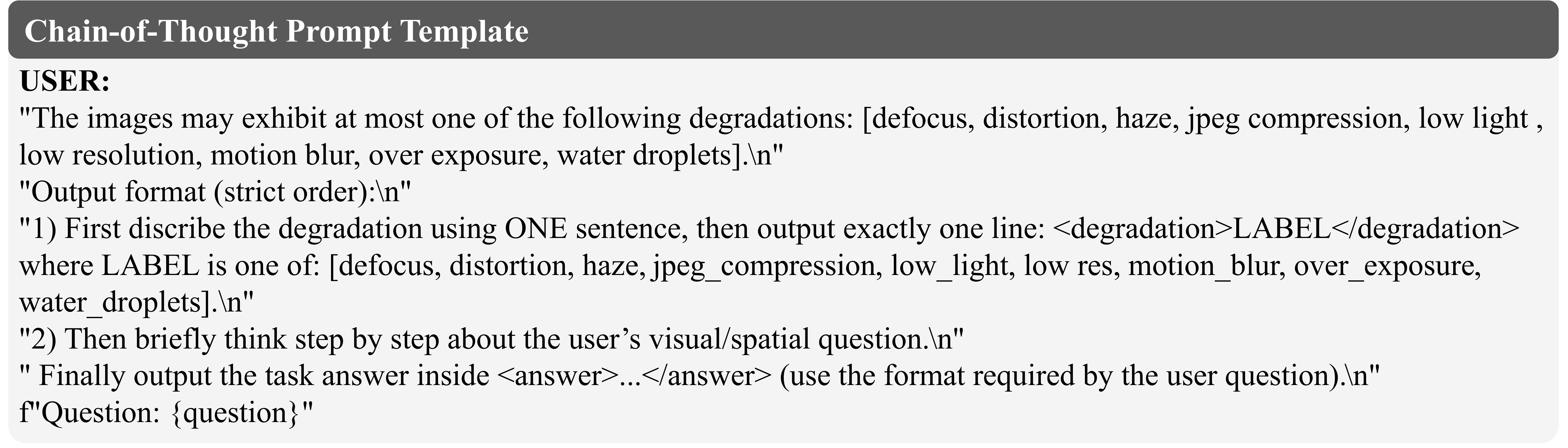}
\caption{Two-stage prompt template for degradation-guided spatial reasoning.}
% \vspace{-0.3cm}
\label{fig:cot_prompt}
\end{figure*}

We provide the two-stage prompt template used to elicit the reasoning capability of Gemini-3.1-Flash-Lite in Figure~\ref{fig:cot_prompt}. The prompt first asks the model to identify the degradation type from a predefined set, and then perform step-by-step spatial reasoning based on the observed degraded images. We further report the degradation recognition accuracy in Table~\ref{tab:acc_dg}. Although Gemini-3.1-Flash-Lite accurately recognizes most degradation types, its performance is substantially lower on haze and low resolution, indicating that degradation attribution itself remains challenging. Such attribution errors propagate to the subsequent reasoning stage and lead to incorrect spatial conclusions.

\begin{table*}[h] 
\centering 
\caption{Accuracy of Gemini-3.1-Flash-Lite for reconizing each degradation.}
\label{tab:acc_dg}
\renewcommand{\arraystretch}{1.1} 
% \vspace{2mm} 
\resizebox{\textwidth}{!}{ 
\begin{tabular}{l ccccccccc} 
\toprule 
\multirow{2}{*}{\textbf{Models}} & \multicolumn{9}{c}{\textbf{Accuracy of Degradation Recognition}}\\ \cmidrule(lr){2-10} 
& Defocus & Distortion & Haze & JPEG-com. & Low-light & Low-res. & Motion-blur & Over-exp. & Water-droplets \\ 
\midrule 
Gemini-3.1-Flash-Lite & 96.6 & 99.9 & 20.4 & 98.0 & 100.0 & 46.9 & 99.7 & 86.2 & 100.0 \\
\bottomrule 
\end{tabular} 
}
\end{table*}

\section{Detailed Degradation Synthesis Pipeline}
\label{sec:appendix_degradation}

\subsection{Optical and Dynamic Degradations}
These degradations are strictly coupled with camera physics and motion. By implementing them internally within the 3DGS rasterizer, we ensure strict geometric consistency across multiple views.

\paragraph{Defocus.} We model the depth-of-field effect caused by a finite camera aperture using the thin-lens approximation. The Circle of Confusion (CoC) radius $r_{\text{CoC}}$ is computed based on the rendered view depth $d$, the focus depth $f$, and the aperture size $a$:
\begin{equation}
    r_{\text{CoC}} = a \frac{|d - f|}{d}
\end{equation}
To simulate this directly within the 3DGS pipeline, this variance is added to the 2D projected covariance matrix of each Gaussian:
\begin{equation}
    \tilde{\boldsymbol{\Sigma}}_{2D} = \boldsymbol{\Sigma}_{2D} + r_{\text{CoC}}^2 \mathbf{I}
\end{equation}
Furthermore, an opacity compensation term $\alpha_{\text{comp}} = \sqrt{\det(\boldsymbol{\Sigma}_{2D}) / \det(\tilde{\boldsymbol{\Sigma}}_{2D})}$ is applied to strictly ensure energy conservation during the differentiable rasterization process.

\paragraph{Distortion.} Real-world wide-angle or fisheye lenses introduce significant non-linear geometric warping. We modify the standard pinhole projection model in the CUDA rasterizer using an equidistant polynomial model. The distorted angle $\theta_d$ is given by:
\begin{equation}
    \theta_d = \theta \left(1 + \sum_{i=1}^4 k_i \theta^{2i}\right)
\end{equation}
where $k_i$ represents the radial distortion coefficients. The Jacobian matrix $\mathbf{J}_{\text{fisheye}}$ is then recomputed via automatic differentiation to accurately project the 3D covariance into the distorted screen space.

\paragraph{Motion Blur.} Caused by camera movement during exposure, motion blur is simulated via the continuous time integration of linear light. We interpolate the camera poses using Spherical Linear Interpolation (Slerp) and accumulate frames over $N$ sub-steps:
\begin{equation}
    I_{\text{blur}} = \frac{1}{N} \sum_{i=0}^{N-1} I_{\text{lin}}(\mathbf{T}(t_i))
\end{equation}
where $\mathbf{T}(t_i)$ denotes the camera extrinsic matrix at time $t_i$ and $N=80$. This approach guarantees highly realistic directional and rotational blur that strictly adheres to the 3D scene geometry.

\subsection{Meteorological Degradations}
These effects depend heavily on the continuous spatial depth of the scene, which is natively provided by our 3DGS representations.

\paragraph{Haze.} We simulate atmospheric scattering using the classic Koschmieder's law. Utilizing the accurate depth map $d(x)$ rendered directly from the 3DGS model, the degraded image intensity $I(x)$ at pixel $x$ is formulated as:
\begin{equation}
    I(x) = J(x)e^{-\beta d(x)} + A\left(1 - e^{-\beta d(x)}\right)
\end{equation}
where $J(x)$ is the original scene radiance, $\beta$ is the scattering coefficient determining the haze density, and $A$ is the global atmospheric light. This model naturally enforces depth-dependent visibility decay.

\paragraph{Water Droplets.} To simulate droplets on the camera lens, we generate a procedural multi-layer height map $h(x, y)$ to derive pixel-wise surface normals $\hat{\mathbf{n}} = (n_x, n_y, n_z)$. These normals are utilized to compute localized refraction offsets based on simplified Snell's law:
\begin{equation}
    \Delta u \propto n_x, \quad \Delta v \propto n_y
\end{equation}
These offsets are combined with local optical blurring and Phong specular highlights to comprehensively mimic the complex optical behavior of water droplets interacting with the scene's light field.

\subsection{Photometric Degradations}
These simulate real-world illumination changes and sensor imperfections. Crucially, these operations are performed in the linear light domain $I_{\text{lin}} = I_{\text{sRGB}}^\gamma$ before final image encoding.

\paragraph{Low-light.} We first scale the scene illumination by an exposure coefficient $e \ll 1$. To simulate the degraded signal-to-noise ratio (SNR) in low-light environments, we inject a physics-based sensor noise model. The noisy observation $I_{\text{noisy}}$ incorporates both photon shot noise (modeled as a Poisson distribution) and read noise (modeled as a Gaussian/Tukey-Lambda distribution):
\begin{equation}
    I_{\text{noisy}} \sim \mathcal{P}\left(\frac{e \cdot I_{\text{lin}}}{k}\right) \cdot k + \mathcal{N}(0, \sigma_{\text{read}}^2)
\end{equation}
where $k$ is the system gain and $\sigma_{\text{read}}$ represents the standard deviation of the electronic read noise.

\paragraph{Over-exposure.} To simulate sensor saturation caused by intense light sources or prolonged exposure, we apply a large exposure gain $e \gg 1$ and inject standard sensor noise. The values are strictly clipped to the sensor's maximum capacity:
\begin{equation}
    I_{\text{clip}} = \max(\min(I_{\text{noisy}}, 1), 0)
\end{equation}
This clipping process is applied prior to Gamma encoding, accurately replicating the irreversible loss of high-frequency textures and geometric details in saturated regions (e.g., near windows or light bulbs).

\subsection{Digital Degradations}
These degradations model common artifacts introduced during post-capture signal processing, storage, and transmission phases. Unlike physical degradations, they operate directly on the encoded 2D image matrix.

\paragraph{JPEG Compression:} We explicitly apply Discrete Cosine Transform (DCT) block quantization controlled by a quality factor $q$. This transformation discards high-frequency coefficients, intentionally introducing the ringing and blocking artifacts typical of low-bandwidth network transmission.
\paragraph{Low-resolution:} We simulate limited sensor resolution or aggressive downsampling by reducing the image resolution of cameras with a scale factor $s$. The image is then upsampled to original resolution. This systematically truncates high-frequency spatial details while maintaining the original image dimensions for the MLLM input format.

\section{Detailed QA Initialization Pipeline}
\label{sec:appendix_qa_generation}
The QA initialization from reconstructed scenes and instances follows Holi-Spatial~\cite{gao2026holispatialevolvingvideostreams}, and we also introducing additional single-view questions, including bbox extent, single-view object distance, object counting and existence.

\paragraph{Instance descriptions.}
For each reconstructed 3D instance, we select its highest-confidence SAM3 mask and overlay the mask contour on the corresponding RGB frame. A VLM is prompted to produce a concise description based on intrinsic, view-stable cues, such as color, material, texture, subtype, text markings, and distinctive structural details. View-dependent phrases are explicitly prohibited, since they would become invalid under a different camera pose. The generated description is stored with the 3D instance and later used as a natural-language reference in QA templates.

\paragraph{View sampling.}
For multi-view questions, we sample image pairs from the scene covisibility matrix while enforcing non-trivial camera motion. The matrix stores pairwise covisibility between every two images in a scene. We compute it in the same manner as MapAnything~\cite{keetha2026mapanything}: depth pixels in a source view are lifted to 3D, reprojected into a target view using the calibrated poses, and counted as covisible only when the target-view depth agrees with the expected reprojected depth under a depth-association threshold. The final score is the normalized number of consistent reprojected pixels. Camera-centric translation questions use a minimum translation baseline, rotation questions use a minimum relative rotation, and object-centric questions require the referenced instance or instances to be visible in the required view(s). For relational questions involving three objects, we additionally require the union of the two views to contain at least three distinct instances and avoid cases where all objects are simultaneously visible in a single image, preventing the task from collapsing into a single-view problem.

\paragraph{Question families.}
SpaceDG instantiates camera-centric, object-centric, and camera-object relational templates. Camera-centric templates cover dominant translation direction, metric translation distance, thresholded translation decisions, and relative rotation. Object-centric templates cover object depth, view-relative direction, inter-object 3D distance, and size or height comparison. Cross-view relational templates ask models to transfer an assumed direction or infer relative position across two views. All ground-truth answers are computed from calibrated camera extrinsics and 3D instance annotations rather than inferred from image pixels.

\paragraph{Option construction and boundary control.}
We use task-specific rules to construct multiple-choice options while avoiding geometrically ambiguous negatives. For camera- and object-direction questions, a secondary axis is included only when its magnitude is sufficiently large relative to the dominant axis ($0.5774\times$, corresponding to a $30^\circ$ angular ratio), which prevents weak off-axis components from changing the textual direction label. For 8-way relative-position questions, the horizontal plane is divided into eight $45^\circ$ sectors. If the ground-truth yaw falls within $3^\circ$ of a sector boundary, the adjacent sector on the boundary side is forbidden as a distractor, so a near-boundary ground truth is not paired with an almost-correct neighboring option. Direction questions whose horizontal displacement is too small, or whose vertical component dominates the horizontal displacement, are discarded.

\paragraph{Geometric ambiguity filters.}
We discard spatial configurations that make the intended relation ill-defined. For triplet-based relation questions, any triplet with intersecting 3D bounding boxes is removed before constructing the local coordinate frame. The local frame is defined by an anchor object and a reference object; if the two centers are too close or the forward direction is nearly collinear with the world-up reference, the sample is skipped. Camera rotation questions are also filtered with a grey-zone rule: relative rotations below $5^\circ$ are discarded, and cases near the boundary between ``single-dominant'' and ``dual-dominant'' rotation are skipped using a $0.1$ buffer around the component-ratio threshold. For thresholded camera-translation questions, the sampled decision threshold is forced away from near equality by moving factors in $(0.85, 1.15)$ to the boundary, reducing accidental ambiguity between ``yes'' and ``no''. For size-comparison questions, objects are labeled as the same length or height only when their computed values differ by less than $10^{-2}$.

\paragraph{Ambiguity filtering.}
Language descriptions can be ambiguous when multiple instances of the same category appear in one image. We therefore build an image-level index over same-label instances. If two same-label instances share the same description, the corresponding QA is discarded. Otherwise, a VLM judge is asked whether the target description uniquely identifies exactly one instance among same-category alternatives; uncertain cases are conservatively dropped. This automated filtering is followed by the manual verification procedure described in Section~\ref{sec:spacedg_dataset_bench}.

\section{Quality Verification}
\subsection{MLLM Filter Prompt Template}
\begin{figure*}[h]
\centering
\includegraphics[width=1\linewidth]{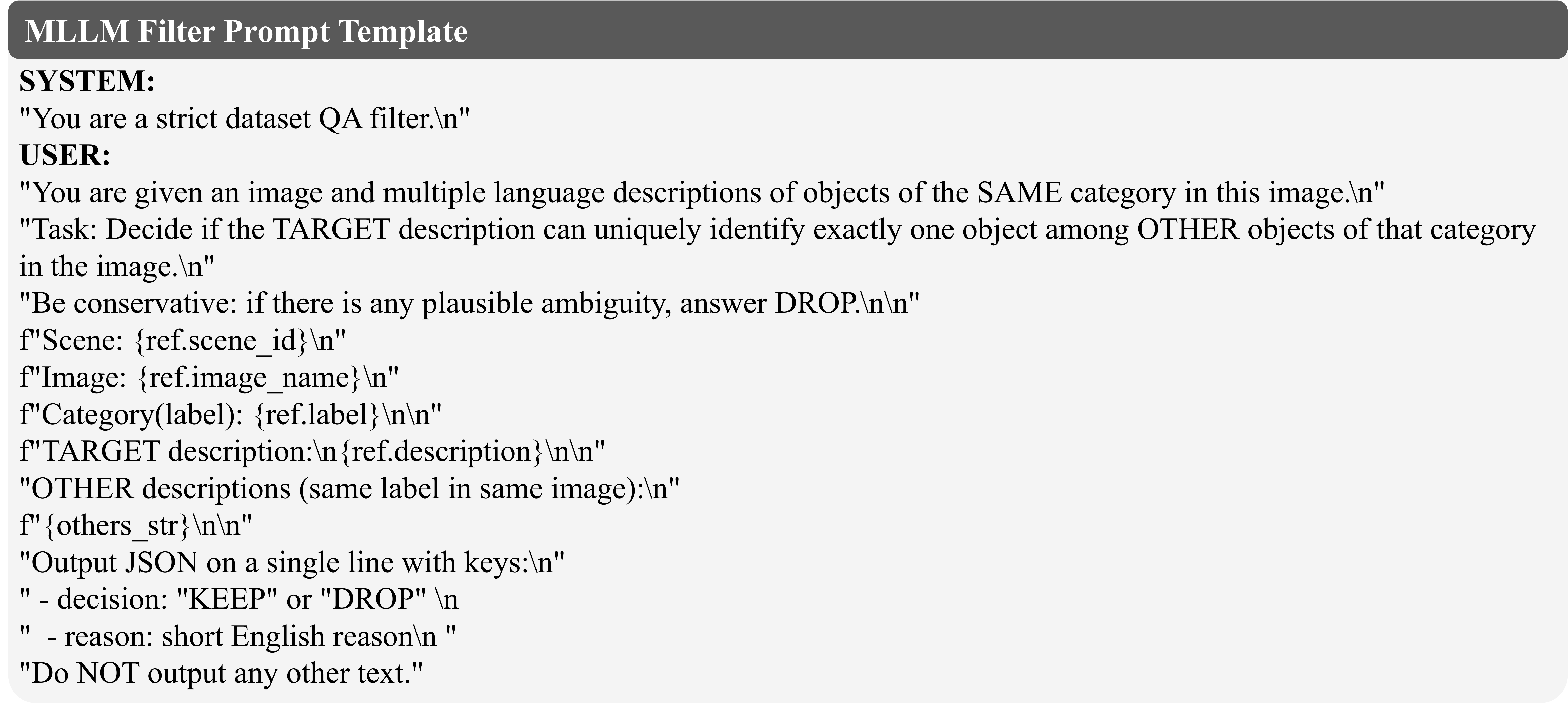}
\caption{MLLM filter prompt template.}
% \vspace{-0.3cm}
\label{fig:filter_prompt}
\end{figure*}

The MLLM filter prompt template is shown in Figure~\ref{fig:filter_prompt}. We provide the MLLM filter with: clean image, segmentation category and generated description. The prompt requires the MLLM to answer with a key word ``KEEP'' or ``DROP'' according to observed ambiguity.

\subsection{Human Review Interface}
\begin{figure*}[h]
\vspace{2mm}
\centering
\includegraphics[width=0.97\linewidth]{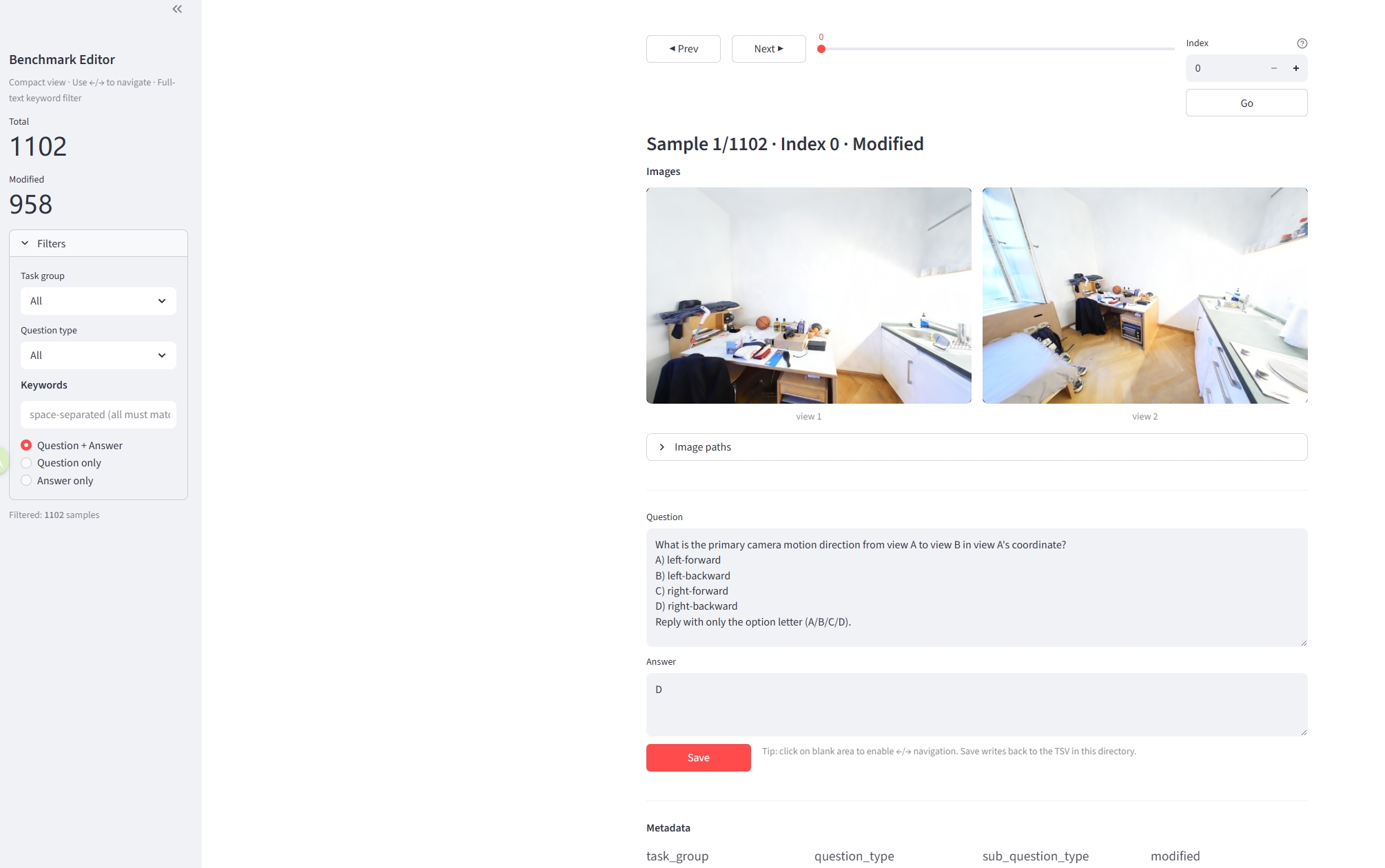}
\caption{Human review interface for SpaceDG-Bench.}
% \vspace{-0.3cm}
\label{fig:interface}
\end{figure*}

We design a benchmark editor interface to support manual verification of SpaceDG-Bench. As shown in the Figure~\ref{fig:interface}, each sample is presented with its rendered image views, the corresponding spatial question, the ground-truth answer, and metadata such as task group and question type. Human reviewers can navigate through samples, filter cases by task or keyword, and directly edit the question or answer when ambiguity, formatting issues, or incorrect labels are observed. All modifications are saved back to the new benchmark file, enabling an efficient and traceable review process for improving the quality and consistency of the final evaluation set.

\section{Question Templates}
\label{app:question-templates}
In this section, we provide detailed tamplates of 11 question types, covering camera-centric, object-centric and camera-object questions.

\subsection{Camera Translation (\texttt{question\_type=camera\_translation})}
\begin{itemize}
\item \textbf{MCQ (main direction).} What is the primary camera motion direction from view A to view B in view A's coordinate?\\
\texttt{<options: A/B/C/D>}\\
Reply with only the option letter (A/B/C/D).
\item \textbf{Numeric (meters).} What is the camera translation distance from view A to view B (meters)?
\item \textbf{Binary (threshold).} Based on the images, decide whether the camera translation from view A to view B exceeds \texttt{<THRESHOLD>} meters. Return ONLY one token: \texttt{'yes'} or \texttt{'no'}.\\
Output format: \texttt{<answer>yes</answer>} or \texttt{<answer>no</answer>} (no extra text).
\end{itemize}

\subsection{Camera Rotation (\texttt{question\_type=camera\_rotation})}
\begin{itemize}
\item \textbf{MCQ (single-axis dominant).} Given view A and view B, consider the relative rotation from A to B expressed in view A's camera frame. Which SINGLE rotation direction is the most prominent?\\
(This question focuses on the dominant axis among \texttt{<axis1>} and \texttt{<axis2>}. )\\
\texttt{<options: A/B/C/D>}
\item \textbf{MCQ (two-axis).} Given view A and view B, consider the relative rotation from A to B expressed in view A's camera frame. Which option best describes the rotation direction using TWO components: \texttt{<axis1>} and \texttt{<axis2>}?\\
\texttt{<options: A/B/C/D>}
\end{itemize}

\subsection{Camera-Object Relative Distance  (\texttt{question\_type=camera\_object\_distance\_estimation})}
\begin{itemize}
\item What is the straight-line distance to the \texttt{\{target\_obj\}} from the camera in meters?
\item How far is the \texttt{\{target\_obj\}} from the current viewpoint?
\item Estimate the physical distance between the camera and the \texttt{\{target\_obj\}}.
\item Can you estimate the straight-line distance to the \texttt{\{target\_obj\}} from the current viewpoint?
\item \textbf{Two-view marked target.} \texttt{<marker text about the target in image A>}\\
Locate the same physical object in image B. Estimate the 3D metric distance (in meters) from the camera position of image B (camera center) to the \texttt{``<label>''} (to the object surface/center point).\\
This is NOT pixel distance. Return only one number in meters (e.g., $0.7$). Output format: \texttt{<answer>NUMBER</answer>}.
\end{itemize}

\subsection{Camera-Object Relative Direction (\texttt{question\_type=camera\_object\_relative\_direction})}
\begin{itemize}
\item What is the relative position of the \texttt{\{target\_obj\_B\}} with respect to the \texttt{\{target\_obj\_A\}} in this view?
\item Where is the \texttt{\{target\_obj\_B\}} located relative to the \texttt{\{target\_obj\_A\}}?
\item Describe the spatial relationship between the \texttt{\{target\_obj\_B\}} and the \texttt{\{target\_obj\_A\}}.
\item In which direction is the \texttt{\{target\_obj\_B\}} compared to the \texttt{\{target\_obj\_A\}}?
\item \textbf{Two-view direction MCQ.} \texttt{<marker text about the target in image A>}\\
Which direction is the \texttt{``<label>''} relative to you when taking image B?\\
\texttt{<options: A/B/C/D>}
\item \textbf{Two-view direction MCQ.} \texttt{<marker text about the target in image A>}\\
When you were taking the photo in Image B, where is the \texttt{<label>} area relative to you?\\
\texttt{<options: A/B/C/D>}
\item \textbf{Two-view relpos MCQ.} \texttt{<marker text for A/B/C across image A and/or B>}\\
You are positioned at \texttt{<labelA>} and face \texttt{<labelB>}. In which direction is \texttt{<labelC>} relative to you?\\
\texttt{<options: A/B/C/D>}
\end{itemize}

\subsection{Camera-Object Cardinal Direction (\texttt{question\_type=cross\_view\_cardinal\_direction})}
\begin{itemize}
\item \textbf{Two-view assumed direction.} \texttt{<marker text for object 1 in image A and object 2 in image B>}\\
The direction of \texttt{<label1>} relative to image A is \texttt{<assumed\_dir>}. What is the direction of \texttt{<label2>} relative to image B?\\
\texttt{<options: A/B/C/D>}
\end{itemize}

\subsection{Object-Object Distance (\texttt{question\_type=inter\_object\_distance})}
\begin{itemize}
\item \textbf{Two-view numeric.} \texttt{<marker text for object 1 in image A and object 2 in image B>}\\
Estimate the 3D metric distance (in meters) between the centers of these two physical objects.\\
Return only one number in meters (e.g., $1.2$). Output format: \texttt{<answer>NUMBER</answer>}.
\end{itemize}

\subsection{Object-Object Cardinal Direction (\texttt{question\_type=object\_proxy\_cardinal\_direction})}
\begin{itemize}
\item \textbf{Proxy-frame MCQ.} \texttt{<marker text for A/B/C across image A and/or B>}\\
The direction of \texttt{<labelA>} relative to \texttt{<labelB>} is \texttt{<assumed\_dir>}. What is the direction of \texttt{<labelC>} relative to \texttt{<labelB>}?\\
\texttt{<options: A/B/C/D>}
\end{itemize}

\subsection{Object Size Comparison (\texttt{question\_type=object\_size\_comparison})}
\begin{itemize}
\item \textbf{MCQ (length).} \texttt{<marker text for object 1 in image A and object 2 in image B>}\\
Which is longer (consider the longest side of the object)? \texttt{<label1>} or \texttt{<label2>}?\\
\texttt{<options: A/B/C/D>}
\item \textbf{MCQ (height).} \texttt{<marker text for object 1 in image A and object 2 in image B>}\\
Which object is taller (consider the top of the objects)? \texttt{<label1>} or \texttt{<label2>}?\\
\texttt{<options: A/B/C/D>}
\end{itemize}

\subsection{Object Bounding-box Size Estimation (\texttt{question\_type=object\_bounding\_size\_estimation})}
\begin{itemize}
\item What are the 3D physical dimensions of the \texttt{\{target\_obj\}}? Please answer in the format \texttt{[shortest edge, middle edge, longest edge]} in meters (e.g. \texttt{[0.10, 0.20, 0.30]}).
\item Estimate the physical size of the \texttt{\{target\_obj\}} in meters, and respond as \texttt{[shortest edge, middle edge, longest edge]} (e.g. \texttt{[0.10, 0.20, 0.30]}).
\item Could you provide the three edge lengths of the \texttt{\{target\_obj\}} in the format \texttt{[shortest, middle, longest]} (meters, e.g. \texttt{[0.10, 0.20, 0.30]})?
\item What is the bounding box extent of the \texttt{\{target\_obj\}}? Reply as \texttt{[shortest edge, middle edge, longest edge]} in meters (e.g. \texttt{[0.10, 0.20, 0.30]}).
\end{itemize}

\subsection{Object Existence Estimation (\texttt{question\_type=object\_existence\_estimation})}
\begin{itemize}
\item Is there a \texttt{\{target\_obj\}} visible in this image?
\item Can you find the \texttt{\{target\_obj\}} in the current view?
\item Does the image contain the \texttt{\{target\_obj\}}?
\item Check if the \texttt{\{target\_obj\}} is present in this picture.
\end{itemize}

\subsection{Object Counting (\texttt{question\_type=object\_counting})}
\begin{itemize}
\item How many \texttt{<label>}s are visible in this image?
\item Count the number of \texttt{<label>} objects in the scene.
\item What is the total count of \texttt{<label>}s shown?
\item Tell me how many \texttt{<label>}s exist in the current view.
\end{itemize}

\section{Additional QA Examples}
\label{sec:appendix_examples}
\begin{figure*}[h]
\centering
\includegraphics[width=0.75\linewidth]{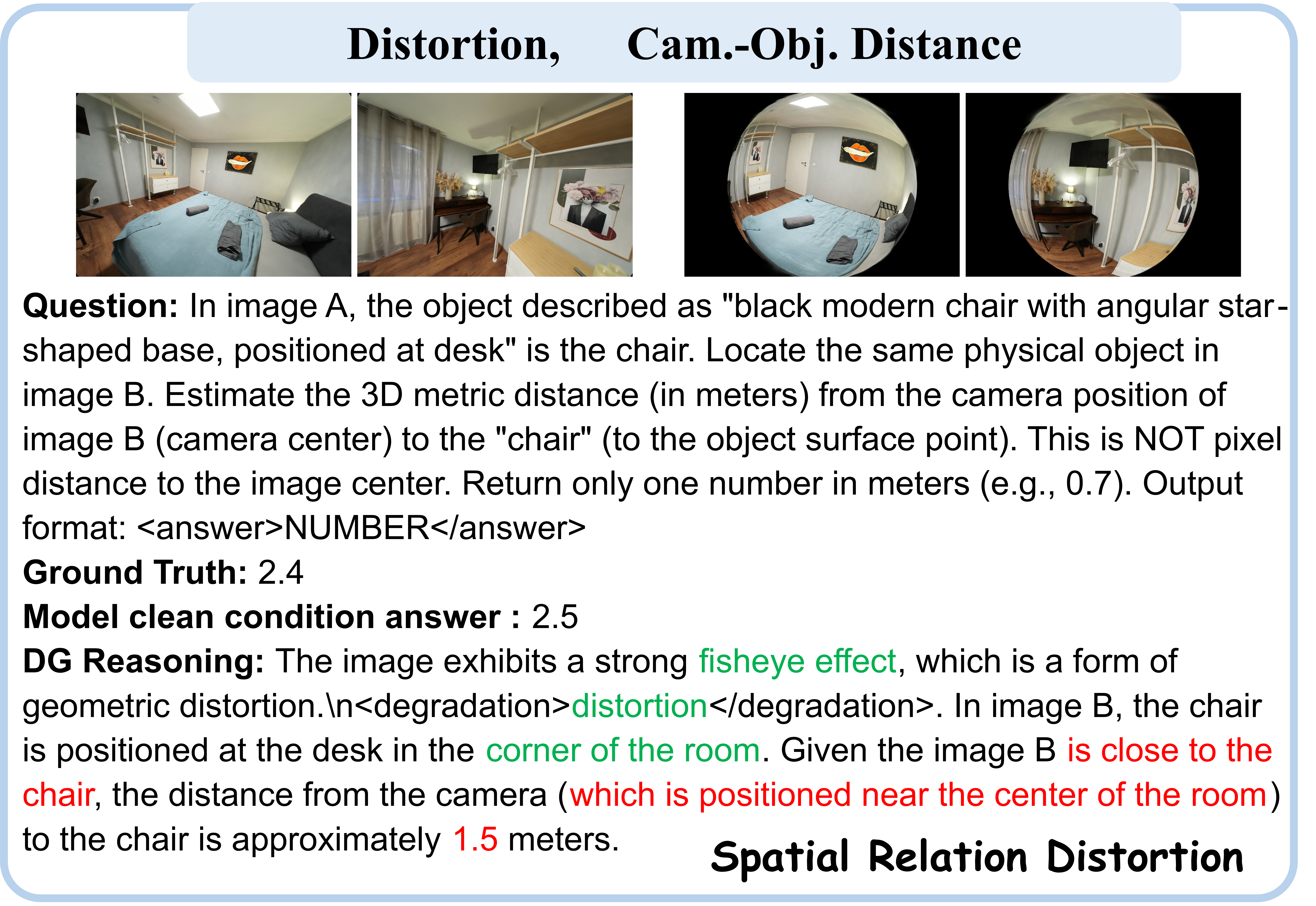}
\caption{Complete QA example of distortion.}
% \vspace{-0.3cm}
\label{fig:distortion}
\end{figure*}

\begin{figure*}[h]
\centering
\includegraphics[width=0.75\linewidth]{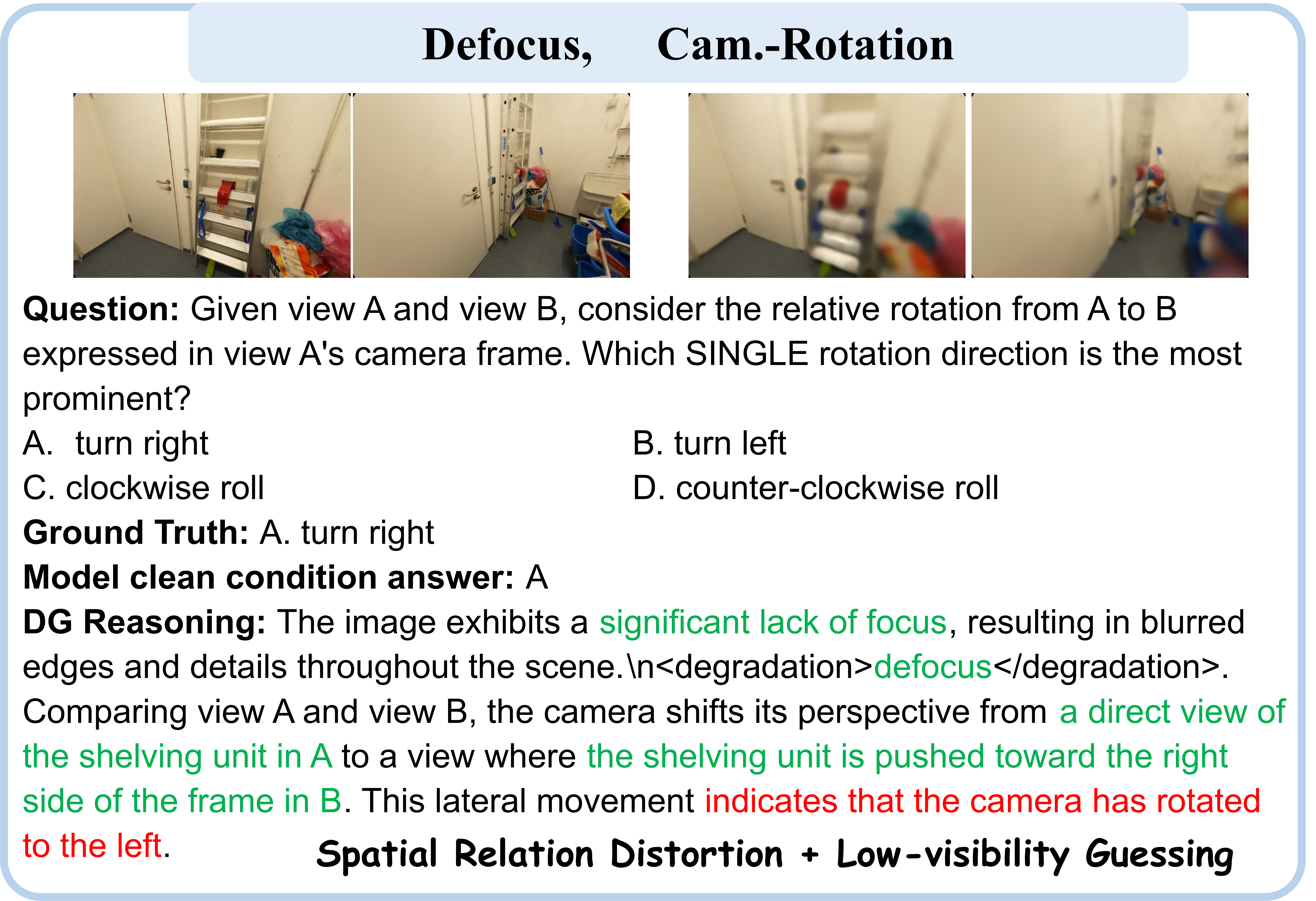}
\caption{Complete QA example of defocus.}
% \vspace{-0.3cm}
\label{fig:defocus}
\end{figure*}

\begin{figure*}[h]
\centering
\includegraphics[width=0.75\linewidth]{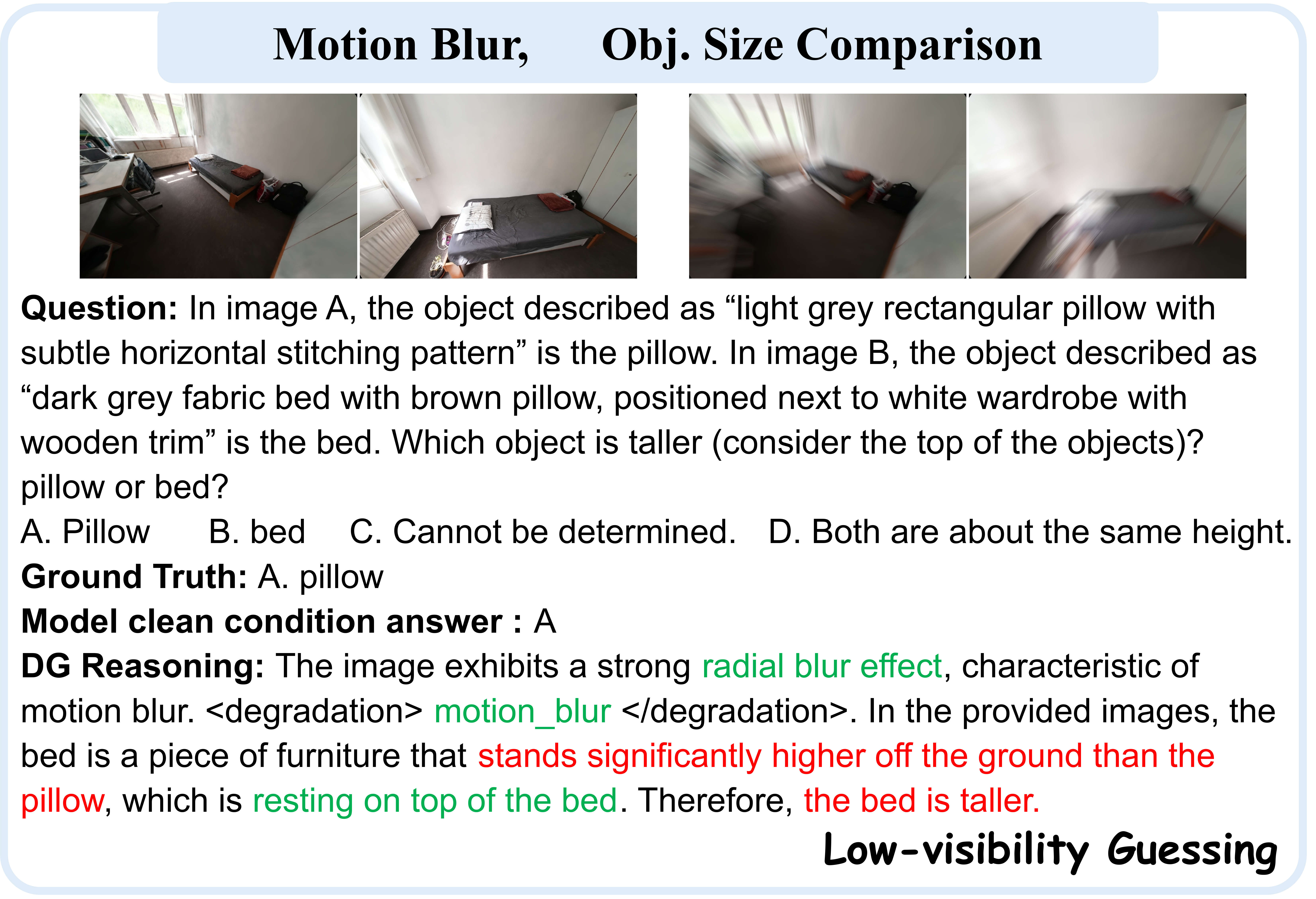}
\caption{Complete QA example of motion blur.}
% \vspace{-0.3cm}
\label{fig:motion}
\end{figure*}

\begin{figure*}[h]
\centering
\includegraphics[width=0.75\linewidth]{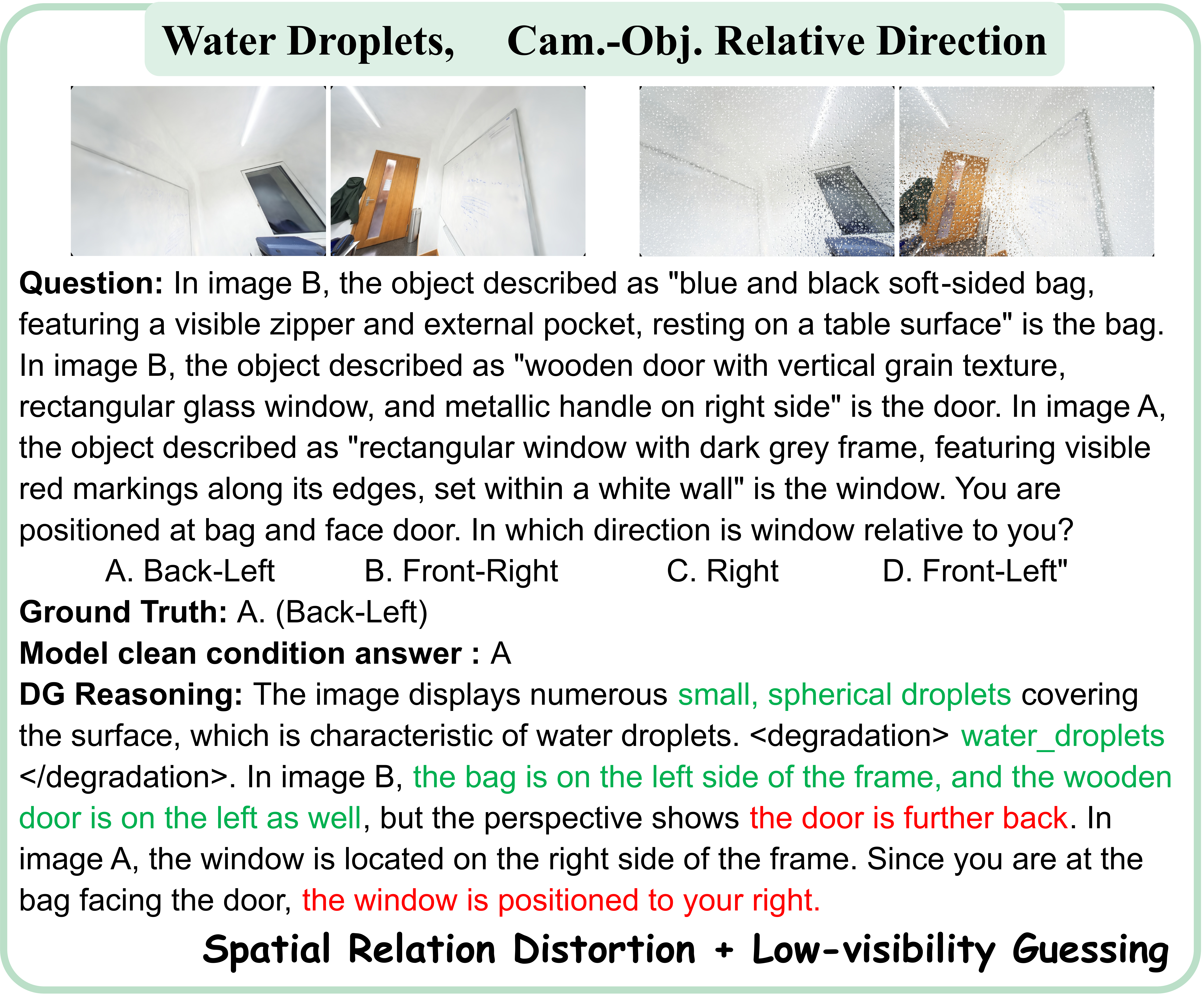}
\caption{Complete QA example of water droplets.}
% \vspace{-0.3cm}
\label{fig:water}
\end{figure*}

\begin{figure*}[h]
\centering
\includegraphics[width=0.75\linewidth]{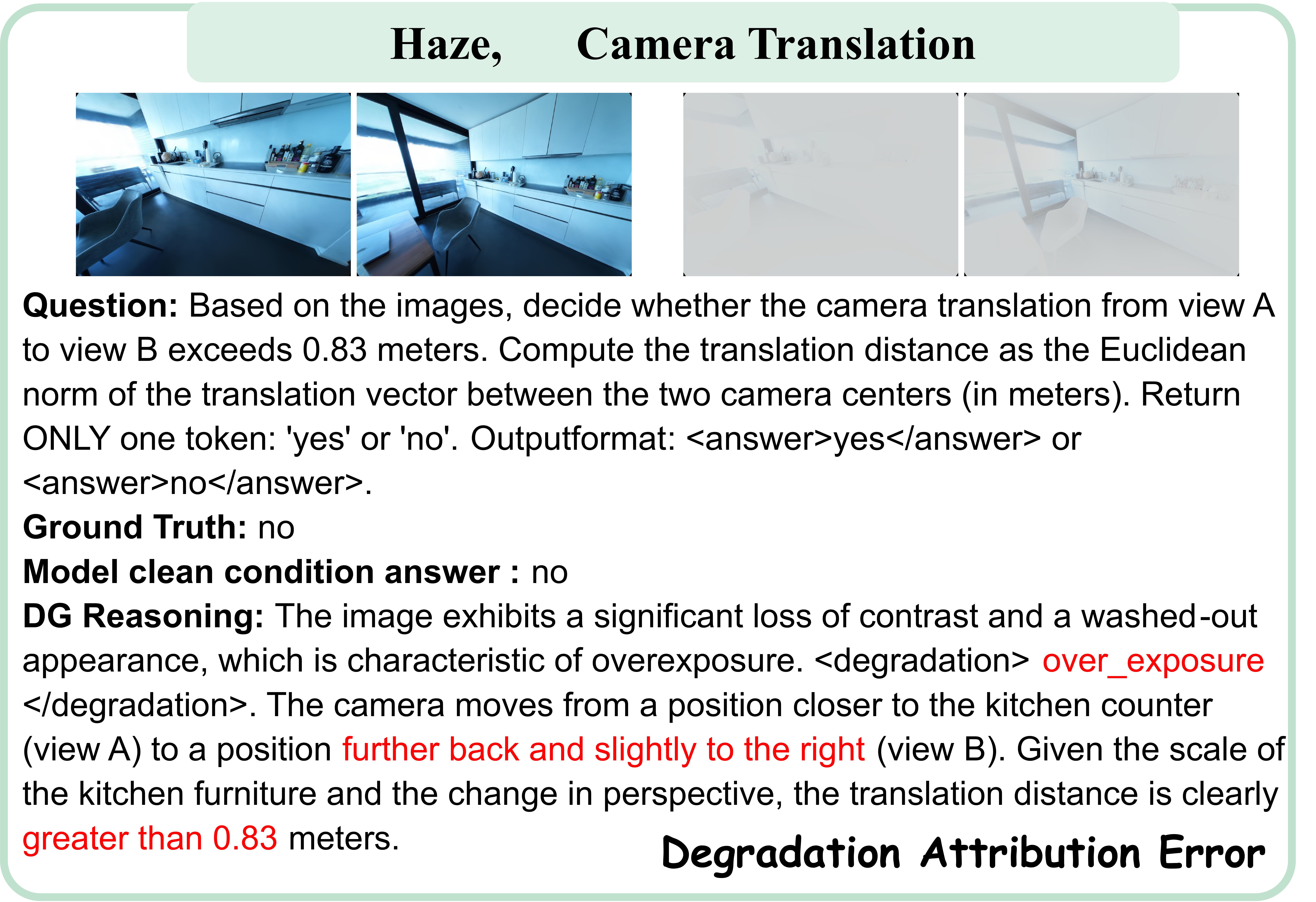}
\caption{Complete QA example of haze.}
% \vspace{-0.3cm}
\label{fig:haze}
\end{figure*}

\begin{figure*}[h]
\centering
\includegraphics[width=0.75\linewidth]{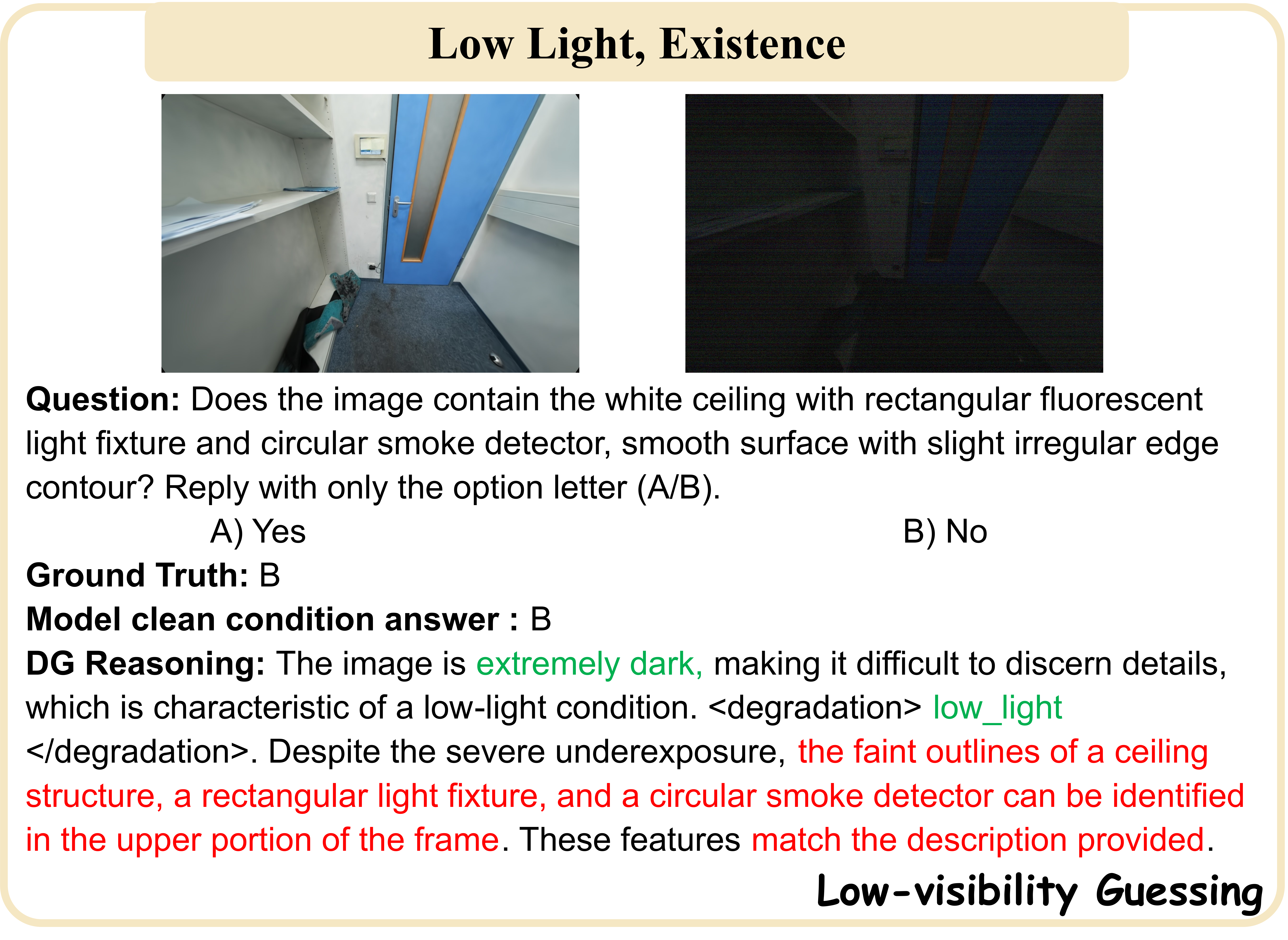}
\caption{Complete QA example of low light.}
% \vspace{-0.3cm}
\label{fig:low_light}
\end{figure*}

\begin{figure*}[h]
\centering
\includegraphics[width=0.75\linewidth]{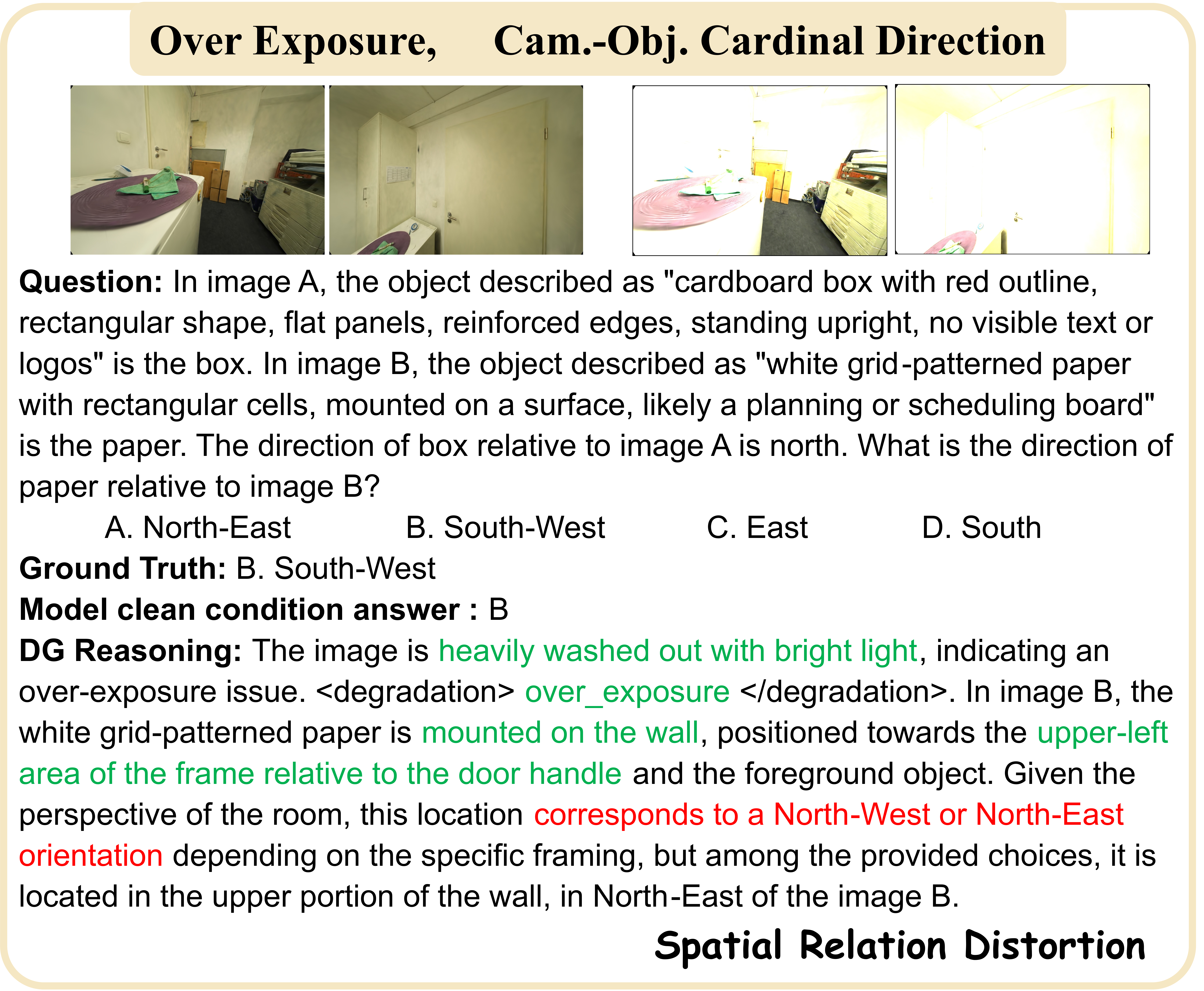}
\caption{Complete QA example of over exposure.}
% \vspace{-0.3cm}
\label{fig:over}
\end{figure*}

\begin{figure*}[h]
\centering
\includegraphics[width=0.75\linewidth]{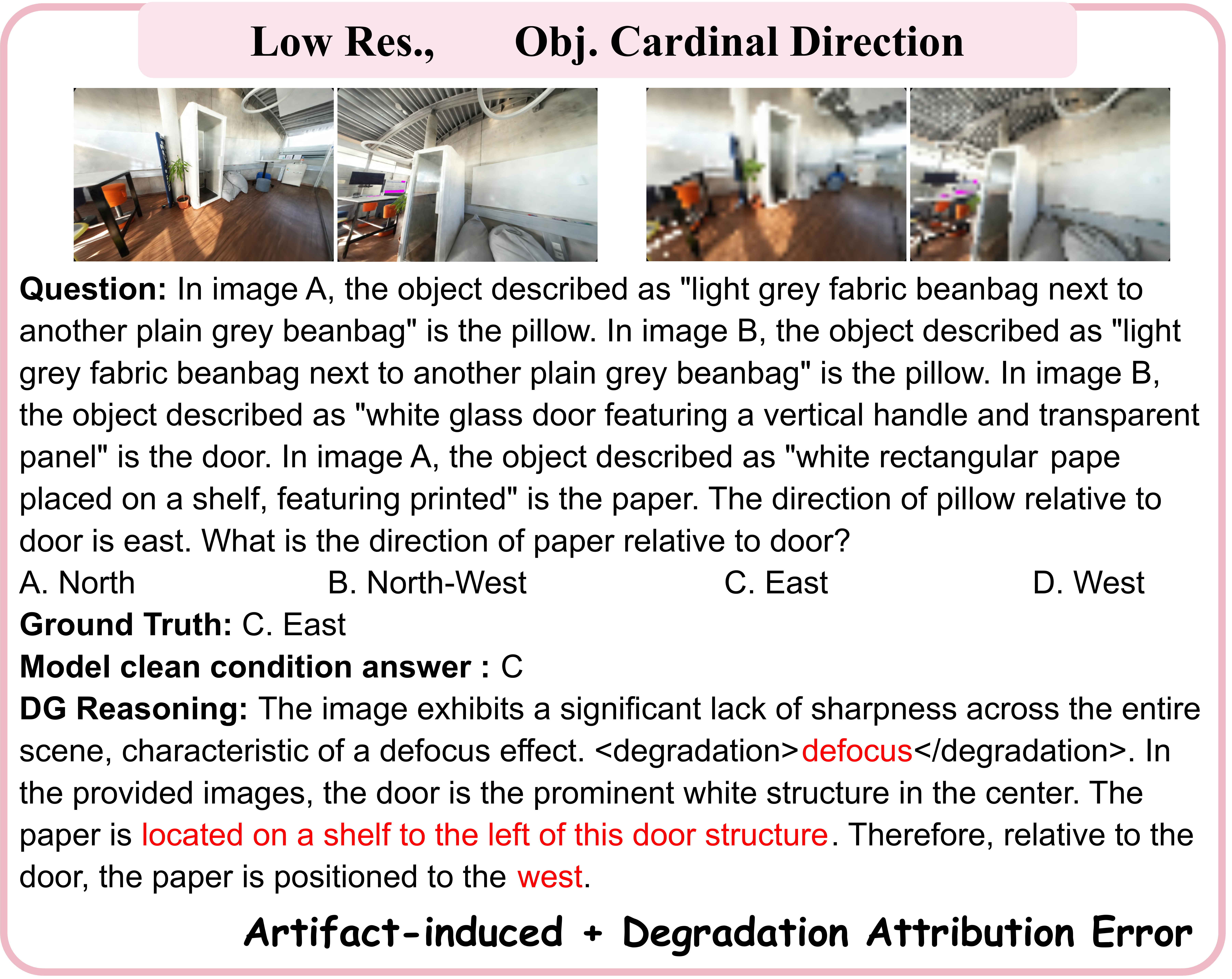}
\caption{Complete QA example of low resolution.}
% \vspace{-0.3cm}
\label{fig:low_res}
\end{figure*}

\begin{figure*}[h]
\centering
\includegraphics[width=0.75\linewidth]{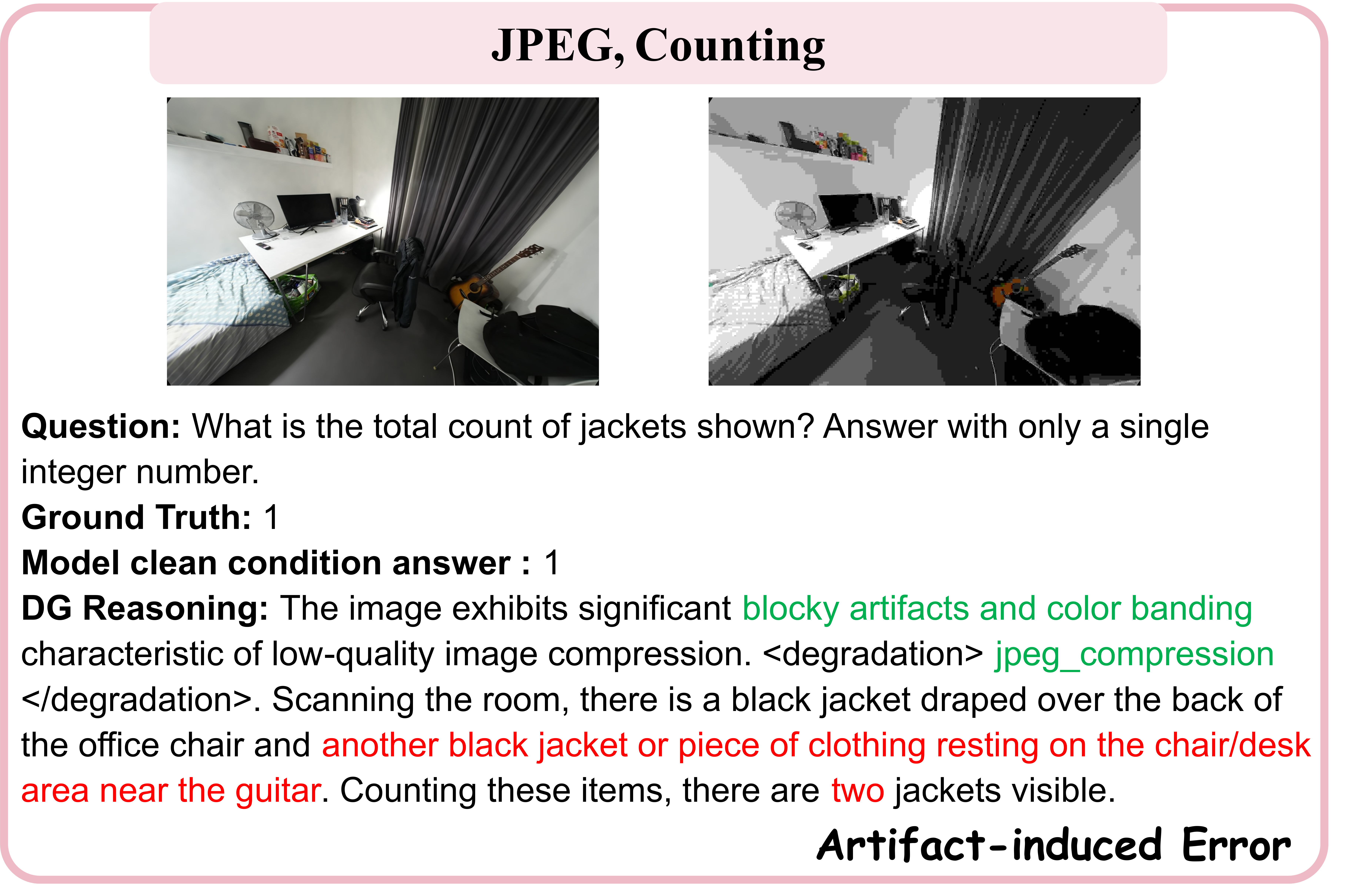}
\caption{Complete QA example of JPEG compression.}
% \vspace{-0.3cm}
\label{fig:jpeg}
\end{figure*}

As shown in Figure~[\ref{fig:distortion}--\ref{fig:jpeg}], we present more error examples of SpaceDG and the reasoning process generated by Gemini-3.1-Flash-Lite. For each example, we provide the complete question, ground truth, model's clean condition answer, and the reasoning process. We label the error type if the model responses a wrong answer.

\section{Limitations}
\label{sec:appendix_limitation}
Despite its large-scale and physically grounded design, SpaceDG has several limitations. First, since the current dataset is built upon ScanNet++, it is primarily restricted to indoor environments. Second, to preserve the physical realism and geometric consistency of synthesized observations, our degradation engine currently supports only nine representative degradation types, leaving other real-world degradations, such as rain, snow, lens flare, rolling-shutter artifacts, and more complex compound corruptions, for future exploration. Nevertheless, SpaceDG provides a systematic and controllable framework for studying degradation-aware spatial intelligence, enabling reliable evaluation and training of MLLMs under realistic visual degradations with accurate 3D spatial ground truth.

\vspace{8mm}

%%%%%%%%%%%%%%%%%%%%%%%%%%%%%%%%%%%%%%%%%%%%%%%%%%%%%%%%%%%%

% \clearpage
% \input{checklist.tex}

\end{document}